\documentclass[10pt,journal,compsoc]{IEEEtran}

\usepackage{cite}
\usepackage{amsmath,amssymb,amsfonts}
\usepackage{graphicx}
\usepackage{textcomp}
\usepackage{xcolor, array, colortbl}

\usepackage[utf8]{inputenc}
\usepackage[T1]{fontenc}      
\usepackage{url}          
\usepackage{booktabs}      
\usepackage{amsfonts}      
\usepackage{nicefrac}    
\usepackage{microtype}     

\usepackage{bbding}
\usepackage{pifont}
\usepackage{amssymb}

\usepackage{bm}
\usepackage{comment}
\usepackage{amsmath}
\usepackage{enumerate} 
\usepackage{enumitem}
\usepackage{mathrsfs}
\usepackage{makeidx}      
\usepackage{graphicx}       
\usepackage{multicol}       
\usepackage{subfigure}
\usepackage{epsfig,amssymb,latexsym}
\usepackage{psfrag}
\usepackage[ruled,vlined]{algorithm2e}
\definecolor{lightgray}{rgb}{.93,.93,.93}

\usepackage{makecell}	
\usepackage{fancyhdr}
\usepackage{multirow}
\usepackage{rotating}
\usepackage{color}

\usepackage[pagebackref=false,breaklinks=true,letterpaper=true,colorlinks,bookmarks=false,linkcolor=red,urlcolor=blue,citecolor=blue]{hyperref}

\newcommand{\mf}{\mathbf}

\newcommand{\mr}{\mathrm}

\usepackage{amsthm}

\ifCLASSINFOpdf
\else
\fi

\begin{document}
\title{No One Left Behind: Real-World Federated Class-Incremental Learning} 

\author{Jiahua Dong, 
        Hongliu Li,
        Yang Cong, \IEEEmembership{Senior Member,~IEEE,}
        Gan Sun,  \\
        Yulun Zhang,
        Luc Van Gool

\IEEEcompsocitemizethanks{
\IEEEcompsocthanksitem Jiahua Dong is with the State Key Laboratory of Robotics, Shenyang Institute of Automation, Chinese Academy of Sciences, Shenyang, 110016, China, also with the Institutes for Robotics and Intelligent Manufacturing, Chinese Academy of Sciences, Shenyang, 110169, China, and also with the University of Chinese Academy of Sciences, Beijing, 100049, China. Email: dongjiahua1995@gmail.com. \protect\\
\vspace{-8pt}
\IEEEcompsocthanksitem Hongliu Li is with the Department of Civil and Environmental Engineering, Hong Kong Polytechnic University, Hong Kong, China. Email: hongliuli1994@gmail.com. \protect\\
\vspace{-8pt}
\IEEEcompsocthanksitem Yang Cong is with the College of Automation Science and Engineering, South China University of Technology, Guangzhou, 510640, China. Email: congyang81@gmail.com. \protect\\ 
\vspace{-8pt}
\IEEEcompsocthanksitem Gan Sun is with the State Key Laboratory of Robotics, Shenyang Institute of Automation, Chinese Academy of Sciences, Shenyang, 110016, China, and also with the Institutes for Robotics and Intelligent Manufacturing, Chinese Academy of Sciences, Shenyang, 110169, China. Email: sungan1412@gmail.com. 
\protect\\
\vspace{-8pt}
\IEEEcompsocthanksitem Yulun Zhang and Luc Van Gool are with the Computer Vision Lab, ETH Z\"{u}rich, Z\"{u}rich 8092, Switzerland. Email: yulun100@gmail.com, vangool@vision.ee.ethz.ch. \protect\\
}

\thanks{Manuscript received April 19, 2005; revised August 26, 2015.}
\thanks{This work was supported in part by the National Nature Science Foundation of China under Grant 62225310, 62127807, 62273333 and 62133005; and the State Key Laboratory of Robotics under Grant 2023-Z13.}
\thanks{The corresponding author is Prof. Yang Cong.}
}

\markboth{IEEE Transactions on Pattern Analysis and Machine Intelligence,~Vol.~14, No.~8, August~2015}
{Shell \MakeLowercase{\textit{et al.}}: Bare Demo of IEEEtran.cls for Computer Society Journals}

\IEEEtitleabstractindextext{
\begin{abstract}
Federated learning (FL) is a hot collaborative training framework via aggregating model parameters of decentralized local clients. However, most FL methods unreasonably assume data categories of FL framework are known and fixed in advance. Moreover, some new local clients that collect novel categories unseen by other clients may be introduced to FL training irregularly. These issues render global model to undergo catastrophic forgetting on old categories, when local clients receive new categories consecutively under limited memory of storing old categories. To tackle the above issues, we propose a novel \underline{L}ocal-\underline{G}lobal \underline{A}nti-forgetting (LGA) model. It ensures no local clients are left behind as they learn new classes continually, by addressing local and global catastrophic forgetting. Specifically, considering tackling class imbalance of local client to surmount local forgetting, we develop a category-balanced gradient-adaptive compensation loss and a category gradient-induced semantic distillation loss. They can balance heterogeneous forgetting speeds of hard-to-forget and easy-to-forget old categories, while ensure consistent class-relations within different tasks. Moreover, a proxy server is designed to tackle global forgetting caused by Non-IID class imbalance between different clients. It augments perturbed prototype images of new categories collected from local clients via self-supervised prototype augmentation, thus improving robustness to choose the best old global model for local-side semantic distillation loss. Experiments on representative datasets verify superior performance of our model against comparison methods. 
The code is available at \url{https://github.com/JiahuaDong/LGA}.

\end{abstract}

}

\maketitle

\IEEEdisplaynontitleabstractindextext
\IEEEpeerreviewmaketitle

\IEEEraisesectionheading{\section{Introduction}\label{sec: introduction}}
\IEEEPARstart{F}{ederated} learning (FL) \cite{DBLP:journals/corr/McMahanMRA16, DBLP:journals/corr/abs-1812-06127, pmlr-v119-karimireddy20a, DBLP:journals/corr/abs-1910-07796} has attracted growing interests in enabling collaborative training across multiple decentralized local clients while providing privacy preservation. Without thoroughly compromising privacy protection of local clients \cite{NEURIPS2020_24389bfe, Qu_2022_CVPR}, it improves the performance of training data-hungry machine learning frameworks via aggregating decentralized local models that are learned on privately-accessible local data \cite{8945292}. Meanwhile, it effectively tackles the data island problem in the real-world \cite{Wang2020Federated} via cooperatively training a global model. Until now, federated learning (FL) \cite{pmlr-v97-yurochkin19a, DBLP:conf/iclr/YangFL21, DBLP:conf/iclr/LiJZKD21} has been successfully applied to a large number of research fields, such as mobile phones \cite{DBLP:journals/corr/abs-1906-04329}, medical diagnosis \cite{liu2021feddg}, intelligent robotics \cite{XIANJIA2021135}, wearable devices \cite{DBLP:journals/corr/abs-1804-07474} and autonomous driving \cite{8917592}.

However, most existing federated learning (FL) models \cite{wang2021addressing, Cheng_2022_CVPR, DBLP:journals/corr/abs-1910-07796, Qu_2022_CVPR, Lu_2022_CVPR} unrealistically assume that the overall FL framework is trained in static application scenarios, where the learned data categories are fixed over time. Obviously, these methods \cite{wang2021addressing, DBLP:journals/corr/abs-1910-07796} cannot be successfully applied to dynamic real-world applications, in which the data of new categories from local clients arrives consecutively under a streaming manner. To tackle this setting, existing FL models \cite{Qu_2022_CVPR, DBLP:journals/corr/abs-1910-07796} require local clients to store all training data of old categories via high cost of memory storage, and then retrain local models to obtain a global class-incremental model via large computation overhead. Unfortunately, it may render the training of FL to be impracticable \cite{DBLP:journals/corr/abs-1906-04329, DBLP:journals/corr/abs-1804-07474} when local clients receive a large number of new categories continuously. Besides, if these models \cite{DBLP:journals/corr/abs-1906-04329, DBLP:journals/corr/abs-1910-07796, DBLP:journals/corr/abs-1804-07474} are required to identify new categories incrementally with limited memory storage to store the training data of old categories, their performance on previously-learned old categories may decrease significantly (\emph{i.e.}, catastrophic forgetting on old categories \cite{Rebuffi_2017_CVPR, Dong_2022_CVPR}). Moreover, some new additional local clients that receive the data of new categories unseen by other local clients may join in the FL training irregularly in real-world applications. The catastrophic forgetting on old categories can be further aggravated by these newly added local clients when training global model via the existing FL framework \cite{pmlr-v97-yurochkin19a, 8945292}.

To tackle above-mentioned practical scenarios \cite{Dong_2022_CVPR}, in this paper, we consider a real-world challenging FL problem, which is referred to as \underline{F}ederated \underline{C}lass-\underline{I}ncremental \underline{L}earning (FCIL). Specifically, for the settings of FCIL, local clients receive the data of new categories incrementally in a streaming manner, and some newly-added local clients collecting unseen novel categories can be introduced to the overall FL training irregularly. More importantly, the current and newly-added local clients have their own preference to receive new categories in an online manner, indicating that class distributions across different local clients are non-independent and identically distributed (Non-IID). These local clients with limited memory available to store training data of old categories \cite{Rebuffi_2017_CVPR, Kang_2022_CVPR, Yan_2021_CVPR, Douillard_2022_CVPR} are required to learn a global class-incremental model collaboratively. This global model can identify new classes continuously under privacy preservation of local clients. Although some local clients only collect a small subset of new categories and store a small subset of old classes, all local clients are required to identify all seen categories, including both old and new classes. In the FCIL settings, no one is left behind when local clients learn new categories consecutively and address forgetting on old classes via collaborative FL training \cite{8945292}.

\begin{figure}[t]
	\centering
	\includegraphics[width=253pt, height=185pt]
	{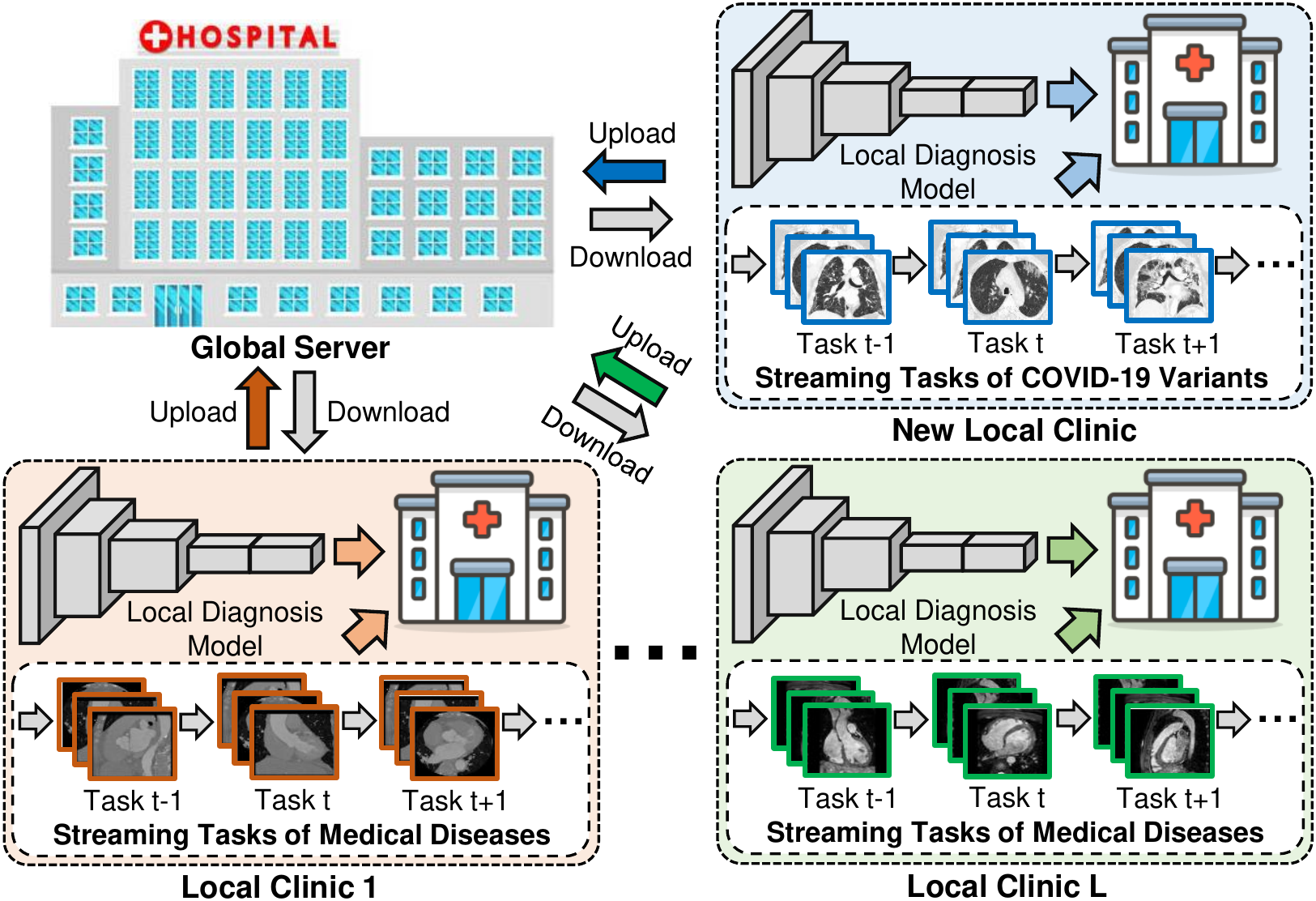}
	\vspace{-20pt}
	\caption{Illustration of our proposed LGA model to address the real-world pandemic COVID-19 diagnosis \cite{dayan2021federated} under the FCIL settings. }
	\label{fig: motivation_of_our_model}
\end{figure}

In this paper, as shown in Fig.~\ref{fig: motivation_of_our_model}, we use pandemic COVID-19 diagnosis \cite{dayan2021federated} as a practical example to better understand the real-world FCIL problem. There are hundreds of hospitals and clinics to learn a global disease diagnosis model collaboratively via the FL training framework \cite{DBLP:journals/corr/abs-1910-07796}, before the pandemic outbreaks. When these hospitals and clinics receive new medical data related to COVID-19, they consider them as new disease category to perform typical FL training \cite{pmlr-v97-yurochkin19a}. Besides, some new hospitals without storing the medical data of old infectious diseases but with collected COVID-19 data may participate in the fight against COVID-19. In the FCIL, all hospitals and clinics cannot be left behind, when they learn to diagnose new COVID-19 variants continuously while alleviate catastrophic forgetting on the old diseases. In this case, existing FL models \cite{pmlr-v119-karimireddy20a, DBLP:journals/corr/abs-1910-07796} suffer from large performance degradation on old diseases when collecting new COVID-19 variants continuously \cite{dayan2021federated}.

To identify new categories consecutively in the FCIL (\emph{e.g.}, diagnosing new COVID-19 variants), a trivial solution is to simply integrate class-incremental learning (CIL) \cite{ICCV2023_HFC, Kim_2023_CVPR} and FL \cite{8945292, DBLP:journals/corr/abs-1804-07474} together. However, it requires global server to have strong prior knowledge about privacy-sensitive information of local clients (\emph{i.e.}, where and when to collect new categories), which heavily compromises privacy preservation in the FL. Moreover, in the FCIL, this solution suffers from local and global catastrophic forgetting on old categories:
\begin{itemize}
\item \textbf{Local catastrophic forgetting} is brought by the class imbalance between old and new categories in each local client. Existing CIL methods \cite{Rebuffi_2017_CVPR, Shmelkov_2017_ICCV, Douillard_2022_CVPR} neglect the large heterogeneity of forgetting speeds among easy-to-forget and hard-to-forget classes within different tasks, which cannot address local forgetting on old categories.  
\item \textbf{Global catastrophic forgetting} indicates the heterogeneous forgetting speeds among different local clients, which is caused by Non-IID class imbalance across local clients. This naive integration solution cannot balance heterogeneous forgetting on old classes among different local clients, further aggravating the local forgetting \cite{Dong_2022_CVPR}. 
\end{itemize}

To address local and global forgetting in the FCIL, in this paper, we develop a novel \underline{L}ocal-\underline{G}lobal \underline{A}nti-forgetting (LGA) model, which is a pioneering exploration to tackle the real-world FCIL problem under the privacy preservation. To be specific, we propose a category-balanced gradient-adaptive compensation loss to overcome local catastrophic forgetting on old categories. It can balance different forgetting speeds of hard-to-forget and easy-to-forget old categories, while normalize the learning speeds of new classes. Besides, a category gradient-induced semantic distillation loss is designed to ensure intrinsic category-relation consistency within different incremental tasks. This loss considers tackling heterogeneous forgetting speeds of old classes when performing semantic relation distillation. Moreover, considering Non-IID class imbalance across clients, we develop a proxy server to surmount global catastrophic forgetting. It significantly improves the distillation gain of the category gradient-induced semantic distillation loss at local side via providing the best old global model from a global perspective. For privacy preservation, we propose a prototype gradient communication strategy to transmit perturbed prototype images of new categories from local clients to proxy server. The proxy server reconstructs these perturbed images and augments them via self-supervised prototype augmentation to pick the best old global model for semantic distillation.

Experiments on representative datasets show significant performance improvements and effectiveness of the proposed LGA model, compared with baseline methods. Several core contributions of this work are summarized as follows: 
\begin{itemize}
\item We propose a real-world challenging FL problem referred to as Federated Class-Incremental Learning (FCIL), where two major challenges are local catastrophic forgetting brought by local clients' class imbalance and global catastrophic forgetting brought by Non-IID class imbalance among different local clients. 
	
\item A novel Local-Global Anti-forgetting (LGA) model is proposed to address FCIL problem via surmounting local and global catastrophic forgetting on old categories. To our best knowledge, our model is the first work to identify new categories consecutively in the FL field. 

\item We tackle local forgetting via developing a category-balanced gradient-adaptive compensation loss and a category gradient-induced semantic distillation loss. They can balance heterogeneous forgetting of hard-to-forget and easy-to-forget old classes, while distill category-relations consistency among different incremental tasks. 

\item A proxy server is designed to collect perturbed prototype images of new classes under privacy preservation via prototype gradient communication. It augments these images via self-supervised prototype augmentation to select the best old model for global anti-forgetting. 
\end{itemize}

This paper is a significant extension of our conference work \cite{Dong_2022_CVPR}. Compared with \cite{Dong_2022_CVPR}, some substantial improvements of this work are listed as follows: 1) We propose a category-balanced gradient-adaptive compensation loss to balance heterogeneous forgetting speeds of hard-to-forget and easy-to-forget old classes via adaptively reweighting category-imbalanced gradient propagation. 2) We design a category gradient-induced semantic distillation loss to distill category-relations consistency among different incremental tasks via considering heterogeneous forgetting speeds of old categories. 3) We ameliorate the proxy server to augment perturbed prototype images via self-supervised prototype augmentation, which can accurately select the best old global model to achieve global anti-forgetting. 4) A large number of qualitative comparison experiments under various FCIL settings and evaluation metrics are conducted on benchmark datasets to further illustrate the effectiveness of our LGA model against baseline comparison methods. 5) More visualization results, ablation studies and insightful analyses are introduced to draw important conclusions.

\section{Related Work}\label{sec: related_work}

\subsection{Federated Learning}
For privacy protection of local clients, federated learning (FL) \cite{zhang2023delving, DBLP:conf/iclr/WangYSPK20, DBLP:conf/iclr/YangFL21, DBLP:conf/iclr/LiJZKD21} focuses on training a decentralized global model via aggregating network parameters of different local models. After McMahan \emph{et al.} \cite{DBLP:journals/corr/McMahanMRA16} propose an average-weighting strategy to aggregate multiple local models when learning global model collaboratively, \cite{DBLP:journals/corr/abs-1812-06127} develops a FedProx framework to address the data heterogeneity for federated local models. Furthermore, \cite{DBLP:journals/corr/abs-1910-07796} encourages local models to learn a shared optimum via introducing a penalty regularizer into objective function. To minimize communication overhead in federated learning, Chen \emph{et al.} \cite{8945292} design a temporal aggregation mechanism via synchronous learning. They aggregate deeper layers in the last few iterations, but integrate shallow layers at every iteration. \cite{pmlr-v97-yurochkin19a, Wang2020Federated} design a Bayesian non-parametric strategy to aggregate model parameters of local clients. 
Yoon \emph{et al.} \cite{Yoon2020FederatedCL} propose federated continual learning (FCL) \cite{zhang2023addressing} via considering a series of continuous tasks within local clients, but they ignore that new local clients collecting novel categories unseen by other clients may be introduced to FL framework. Besides, \cite{Yoon2020FederatedCL} assumes all tasks have the same and fixed class number, while it has strong prior knowledge about task indices for testing. It is significantly different from FCIL problem that can learn varying number of new classes in each task and classify all classes observed so far without prior knowledge about task indices. 
Peng \emph{et al.} \cite{DBLP:conf/iclr/PengHZS20} rely on domain adaptation technology \cite{9616392_Dong, What_Transferred_Dong_CVPR2020} to improve the generalization performance on unsupervised target domain at local side under the FL settings. 
Qu \emph{et al.} \cite{Qu_2022_CVPR} address performance degradation of heterogeneous data across local clients when deploying the FL model on different devices with large distribution shift. 
However, above FL methods \cite{pmlr-v97-yurochkin19a, 8945292, Wang2020Federated} cannot learn new categories consecutively in a streaming manner, and suffer from forgetting on old categories when local memory is limited to store old categories.

\subsection{Class-Incremental Learning}
Class-incremental learning (CIL) \cite{Yang_2023_CVPR, DBLP:conf/aaai/KimC21, Gao_2023_CVPR, 9599446} focuses on identifying new categories continuously in the real-world. We divide the existing CIL methods \cite{10.1007/978-3-030-58565-5_6, 9864267, 9915459} into three categories: learning without access to training data of old classes, generative replay of old classes and exemplar memory construction of old classes. Specifically, when the training data of old classes is unavailable, Kirkpatrick \emph{et al.} \cite{Kirkpatrick3521} propose to compensate biased optimization brought by new classes, and knowledge distillation is employed by  \cite{10.1007/978-3-319-46493-0_37, Shmelkov_2017_ICCV, zhang2022ideal} to tackle performance degradation (\emph{i.e.}, catastrophic forgetting) on old categories. For the generative replay of old classes, \cite{NIPS2018_7836} relies on adversarial learning to design memory replay generator of old classes, and surmounts catastrophic forgetting via performing replay alignment. Besides, \cite{10.5555/3294996.3295059} proposes a dual cooperative model including a memory generator to synthesize old classes and a task solver to address forgetting. As introduced in \cite{Rebuffi_2017_CVPR, wu2019large, 10.1007/978-3-030-58565-5_6}, class-imbalanced distribution between old and new categories is main reason for exemplar memory-based CIL methods \cite{Christian2021MGeoCont, Douillard_2022_CVPR} to suffer from catastrophic forgetting. To be specific, \cite{Hu_2021_CVPR} considers distilling causal relations of class-imbalanced training samples. Simon \emph{et al.} \cite{Christian2021MGeoCont} propose an improved knowledge distillation technology, and utilize geodesic path to measure the similarity between old and new predictions. \cite{Shi_2022_CVPR, Liu2020AANets, Kang_2022_CVPR} design an adaptive network to balance the stability and plasticity. \cite{Douillard_2022_CVPR} introduces transformer framework to tackle forgetting on old classes via designing expandable task tokens. However, to tackle the FCIL problem, these existing CIL methods \cite{Rebuffi_2017_CVPR, Douillard_2022_CVPR} require strong privacy knowledge about where and when to collect new classes, which is impracticable to violate the requirement of privacy preservation in real-world applications.

\section{Problem Definition}\label{sec: problem_definition}

As claimed in traditional class-incremental learning (CIL) \cite{Rebuffi_2017_CVPR, Wu_2022_CVPR, Christian2021MGeoCont, Liu2020AANets, ICCV2023_HFC}, a series of continuous learning tasks are denoted as $\mathcal{T} = \{\mathcal{T}^t\}_{t=1}^T$, where $T$ represents the number of consecutive tasks. For the $t$-th learning task $\mathcal{T}^t = \{\mf{x}_i^t, \mf{y}_i^t\}_{i=1}^{N^t}$, $\mf{x}_i^t$ and $\mf{y}_i^t\in \mathcal{Y}^t$ denote the $i$-th image and its corresponding one-hot label, where $N^t$ is the number of samples in $\mathcal{T}^t$ and $\mathcal{Y}^t$ denotes the label space of $\mathcal{T}^t$ consisting of $C^t$ new categories. There are no overlapped label spaces between any two learning tasks: $\mathcal{Y}^t\cap(\cup_{j=1}^{t-1}\mathcal{Y}^j) = \emptyset$. That is to say, $C^t$ new categories in $\mathcal{T}^t$ are different from $C^o=\sum_{i=1}^{t-1}C^i \subset \cup_{j=1}^{t-1}\mathcal{Y}^j$ old categories learned in previous consecutive tasks $\{\mathcal{T}^i\}_{i=1}^{t-1}$. In the $t$-th task $\mathcal{T}^t$, motivated by \cite{Rebuffi_2017_CVPR, wu2019large, Douillard_2022_CVPR, Kang_2022_CVPR}, we allocate an exemplar memory $\mathcal{M}$ to store $\frac{|\mathcal{M}|}{C^o}$ images per old category, and $\mathcal{M}$ satisfies the constraint $\frac{N^t}{C^t} \gg \frac{|\mathcal{M}|}{C^o}$.

\begin{figure*}[t]
\centering
\includegraphics[width=520pt, height=195pt]
{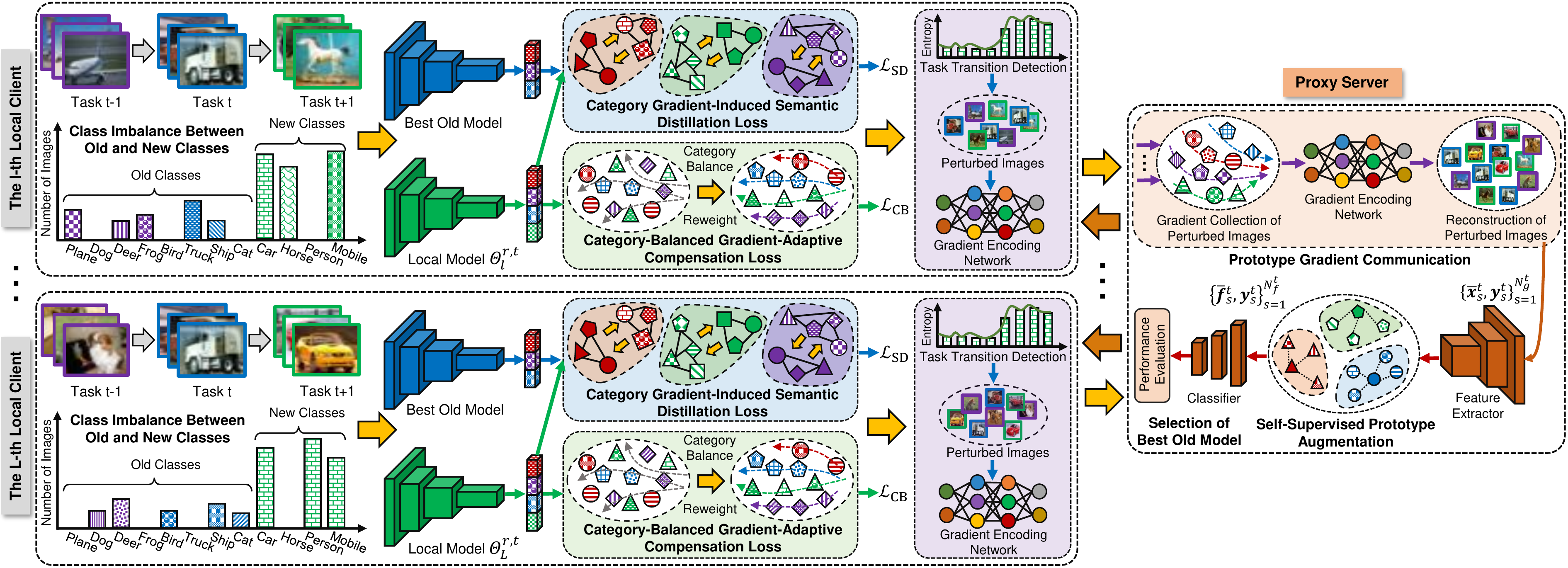}
\vspace{-20pt}
\caption{Overview framework of the proposed LGA model. It can surmount local catastrophic forgetting brought by local clients' class imbalance via a \textit{category-balanced gradient-adaptive compensation loss} $\mathcal{L}_{\mr{CB}}$ and a \textit{category gradient-induced semantic distillation loss} $\mathcal{L}_{\mr{SD}}$, while can also alleviate global catastrophic forgetting that is caused by Non-IID class imbalance among different local clients via a \textit{proxy server} $\mathcal{S}_p$. We develop a prototype gradient communication strategy to privately communicate perturbed prototype images between local clients and $\mathcal{S}_p$, while select the best old global model accurately for semantic distillation loss $\mathcal{L}_{\mr{SD}}$ via performing self-supervised prototype augmentation to augment perturbed images.  }
\label{fig: overview_of_our_model}
\end{figure*}

For the federated class-incremental learning (FCIL), we follow the baseline settings proposed in \cite{Dong_2022_CVPR}, which is extended from traditional class-incremental learning (CIL). Denote a global server as $\mathcal{S}_g$ and $L$ local clients as $\{\mathcal{S}_l\}_{l=1}^L$. We randomly select several local clients from $\{\mathcal{S}_l\}_{l=1}^L$ to perform gradient aggregation at the $r$-th ($r=1,\cdots, R$) global round, where $R$ is total number of global rounds. For the FCIL settings, when the $l$-th local client $\mathcal{S}_l$ is selected for gradient aggregation in the $t$-th incremental task, the global server $\mathcal{S}_g$ will distribute the latest global model $\Theta^{r, t}$ to it. Then the selected $\mathcal{S}_l$ can optimize $\Theta^{r, t}$ on its own private training data $\mathcal{T}_l^t\cup\mathcal{M}_l\sim\mathcal{P}_l^{|\mathcal{T}_l^t|+|\mathcal{M}_l|}$ in the $t$-th incremental task, where $\mathcal{T}_l^t=\{\mf{x}_{li}^t, \mf{y}_{li}^t\}_{i=1}^{N_l^t}\subset \mathcal{T}^t$ represents $N_l^t$ pairs of image $\mf{x}_{li}^t$ and label $\mf{y}_{li}^t$, $\mathcal{M}_l$ is allocated exemplar memory to store some representative images of old classes seen by the $l$-th client $\mathcal{S}_l$, and $\mathcal{P}_l$ is the data distribution of $\mathcal{S}_l$ for the $t$-th incremental task. $\{\mathcal{P}_l\}_{l=1}^L$ are non-independent and identically distributed (Non-IID) across local clients. 
Moreover, the label space $\mathcal{Y}_l^t\subset \mathcal{Y}^t$ of the $l$-th client $\mathcal{S}_l$ is a subset of $\mathcal{Y}^t=\cup_{l=1}^L\mathcal{Y}_l^t$ in the $t$-th incremental task, where $\mathcal{Y}_l^t$ consists of $C_l^t$ ($C_l^t \leq C^t$) new categories that have no overlap with $C_l^o=\sum_{i=1}^{t-1}C_l^i\subset \cup_{j=1}^{t-1}\mathcal{Y}_l^j$ old categories. 
After receiving global model $\Theta^{r, t}$ to perform local training on private data $\mathcal{T}_l^t\cup\mathcal{M}_l$, $S_l$ obtains a locally updated model $\Theta_{l}^{r, t}$ in the $t$-th incremental task. The global server $\mathcal{S}_g$ collects locally updated models of all selected local clients to aggregate them as global model $\Theta^{r+1, t}$ of next global round, which is then distributed to all clients for local training.

In each incremental task, $L$ local clients $\{\mathcal{S}_l\}_{l=1}^L$ are classified into three categories (\emph{i.e.}, \{$\mathcal{S}_l\}_{l=1}^L = {\mathcal{Z}_o}\cup\mathcal{Z}_n\cup \mathcal{Z}_a$). To be specific, $\mathcal{Z}_o$ has $L_o$ local clients with exemplar memory constructed via old learned tasks but without access to the new categories of current task; $\mathcal{Z}_n$ consists of $L_n$ local clients allocating exemplar memory of old categories while receiving new categories of current task; and $\mathcal{Z}_a$ is composed of $L_a$ newly-added clients that have no learning experience of old categories but collect new categories of current task. As introduced in \cite{Dong_2022_CVPR}, we add new local clients $\mathcal{Z}_a$ irregularly at any global round, and randomly select $\{\mathcal{Z}_o, \mathcal{Z}_n, \mathcal{Z}_a\}$ without prior knowledge at each global round. The number of local clients $L=L_o+L_n+L_a$ increases gradually as consecutive learning tasks, due to dynamic change of $\{\mathcal{Z}_o, \mathcal{Z}_n, \mathcal{Z}_a\}$.

In the FCIL settings, we have no human prior about data distributions $\{\mathcal{P}_l\}_{l=1}^L$, the number of continuous learning tasks $T$, when to introduce new clients and receive new categories. To address the FCIL problem, we aim to learn a global model $\Theta^{R, T}$ to identify new categories continuously and tackle the forgetting on old categories under privacy protection of local data, via communicating model parameters between local clients $\{\mathcal{S}_l\}_{l=1}^L$ and global server $\mathcal{S}_g$.

\section{The Proposed Model}\label{sec: the_proposed_model}
The graphical illustration of proposed LGA model is presented in Fig.~\ref{fig: overview_of_our_model}. To tackle the FCIL problem, our LGA model performs local anti-forgetting (Section~\ref{sec: local_anti_forgetting}) via a category-balanced gradient-adaptive compensation loss and a category gradient-induced semantic distillation loss, while achieving global anti-forgetting (Section~\ref{sec: global_anti_forgetting}) via proxy server to improve local distillation gain from a global perspective.

\subsection{Local Anti-Forgetting}\label{sec: local_anti_forgetting}
The $l$-th local client $\mathcal{S}_l\in\mathcal{Z}_n$ with privately accessible exemplar memory $\mathcal{M}_l$ can receive new training data $\mathcal{T}_l^t$ in the $t$-th incremental task. Given a mini-batch $\{\mathbf{X}_{lb}^t, \mathbf{Y}_{lb}^t\} = \{\mf{x}_{li}^t, \mf{y}_{li}^t\}_{i=1}^B\subset \mathcal{T}_l^t\cup\mathcal{M}_l$, the classification loss $\mathcal{L}_{\mr{CE}}$ for the $l$-th client $\mathcal{S}_l$ is defined as follows:
\begin{align}
	\mathcal{L}_{\mr{CE}} =\frac{1}{B}\sum_{i=1}^B \mathcal{D}_{\mr{CE}}(P_l^t(\mathbf{x}_{li}^t, \Theta^{r, t}), \mathbf{y}_{li}^t),
\label{eq: classification_loss}
\end{align}
where $\Theta^{r, t}$ represents the global classification model in the $t$-th incremental task, and its model parameters are distributed from global server $\mathcal{S}_g$ to local clients at the $r$-th global round. $P_l^t(\mathbf{x}_{li}^t, \Theta^{r, t})\in\mathbb{R}^{C^o+C^t}$ is the probability prediction of global model $\Theta^{r, t}$. $B$ represents the batch size, and $\mathcal{D}_{\mr{CE}}(\cdot, \cdot)$ denotes the standard cross-entropy loss.

However, the severe class imbalance between old and new categories (\emph{i.e.}, $\frac{N_l^t}{C_l^t} \gg \frac{|\mathcal{M}_l|}{C_l^o}$) in the $l$-th local client $\mathcal{S}_l$ enforces its local training (optimized via Eq.~\eqref{eq: classification_loss}) to significantly decrease the performance on old categories. That is to say, $\mathcal{S}_l$ suffers from local catastrophic forgetting on old classes when minimizing Eq.~\eqref{eq: classification_loss} for local model training. To achieve local anti-forgetting for local clients, we develop a category-balanced gradient-adaptive compensation loss to balance heterogeneous forgetting speeds of hard-to-forget and easy-to-forget old categories, while propose a category gradient-induced semantic distillation loss to distill consistent category relations within different incremental tasks.

\textbf{$\bullet$ Category-Balanced Gradient-Adaptive Compensation: } 
When local client $\mathcal{S}_l$ receives global model $\Theta^{r, t}$ from global server $\mathcal{S}_g$ at the $r$-th global round, the class imbalance within local client leads to gradient-imbalanced back-propagation of the classifier in $\Theta^{r, t}$. It enforces the training of local model $\Theta_l^{r, t}$ to suffer from various updating speeds for old and new categories (\emph{i.e.}, different forgetting speeds for old categories and different learning speeds for new categories). Such heterogeneous updating speeds among different classes significantly aggravate catastrophic forgetting on old classes, when new training data from continuous learning tasks becomes a subset of old learned categories consecutively.

To achieve this, Dong \emph{et al.} \cite{Dong_2022_CVPR} propose to reweight imbalanced gradient back-propagation of old and new categories. However, it normalizes heterogeneous forgetting speeds of all old categories via an unified gradient normalization mean instead of task-specific gradient means. Such strategy unreasonably assumes all old classes have the same forgetting speed, and neglects significant heterogeneity of forgetting speeds among old categories from different incremental tasks. Thus, the reweighting strategy in \cite{Dong_2022_CVPR} enforces the model to bias towards some hard-to-forget categories learned in recent tasks, while ignoring other easy-to-forget categories from long-ago learned tasks. Besides, the hard-to-forget and easy-to-forget old categories change dynamically as the consecutive learning tasks, further increasing the difficulty to balance heterogeneous forgetting of old categories.

To tackle the above challenges, as depicted in Fig.~\ref{fig: overview_of_our_model}, we propose a category-balanced gradient-adaptive compensation loss $\mathcal{L}_{\mr{CB}}$. It can balance heterogeneous forgetting speeds of hard-to-forget and easy-to-forget old categories, while normalize learning speeds of new categories via considering category-level balance to adaptively reweight gradient back-propagation. Given a sample $(\mf{x}_{li}^t, \mf{y}_{li}^t)\subset\mathcal{T}_l^t\cup\mathcal{M}_l$ from the $t$-th task, as introduced in \cite{wang2021addressing}, gradient updating value $\mathcal{V}_{li}^t$ with respect to the $y_{li}^t$-th ($1\leq y_{li}^t\leq C^o+C^t$) neuron $\mathcal{N}_{y_{li}^t}^t$ of the last layer of classifier in local model $\Theta_l^{r, t}$ is expressed as:
\begin{align}
	\mathcal{V}_{li}^t= \frac{\partial\mathcal{D}_{\mr{CE}}(P_l^t(\mathbf{x}_{li}^t, \Theta_l^{r, t}), \mathbf{y}_{li}^t)}{\partial\mathcal{N}_{y_{li}^t}^t} = P_{l}^t(\mf{x}_{li}^t, \Theta_l^{r, t})_{y_{li}^t} - 1,
	\label{eq: gradient_measurement}
\end{align}
where $y_{li}^t$ is scalar ground-truth label of $\mf{x}_{li}^t$, and $\mf{y}_{li}^t$ denotes its one-hot encoding vector. $P_{l}^t(\mf{x}_{li}^t, \Theta_l^{r, t})_{y_{li}^t}$ is softmax probability of the $y_{li}^t$-th category predicted via local model $\Theta_l^{r, t}$.

To balance different forgetting speeds of hard-to-forget and easy-to-forget old categories while normalizing the learning speeds of new categories, we perform separate gradient normalization for the categories learned by different incremental tasks, and use it to reweight traditional cross-entropy loss $\mathcal{L}_{\mr{CE}}$ for category-balanced gradient compensation. Given a mini-batch $\{\mathbf{x}_{li}^t,\mathbf{y}_{li}^t\}_{i=1}^B$ sampled from the $t$-th task, we define gradient mean $\mathcal{V}_{\kappa}$ of the classes that have been learned in the $\kappa$-th ($1\leq\kappa\leq t$) incremental task as:
\begin{align}
	\mathcal{V}_{\kappa} = \frac{1}{\sum_{i=1}^B \mathbb{I}_{\mf{y}_{li}^t\in\mathcal{Y}_l^\kappa}} \sum\nolimits_{i=1}^B |\mathcal{V}_{li}^t|\cdot \mathbb{I}_{\mf{y}_{li}^t\in\mathcal{Y}_l^{\kappa}}, 
	\label{eq: task_specific_gradient}
\end{align}
where $\mathbb{I}_{(\cdot)}$ represents the indicator function. When its subscript condition is false, ${\mathbb{I}}_{(\mr{False})} = 0$; otherwise, ${\mathbb{I}}_{(\mr{True})} = 1$.

The task-specific gradient means $\{\mathcal{V}_\kappa\}_{\kappa=1}^t$ obtained via Eq.~\eqref{eq: task_specific_gradient} can measure heterogeneous forgetting speeds of old categories from different tasks and learning speeds of new categories. Compared with \cite{Dong_2022_CVPR}, we can achieve category-balanced gradient back-propagation via $\{\mathcal{V}_\kappa\}_{\kappa=1}^t$ rather than an universal gradient mean across all old learning tasks. However, the hard-to-forget and easy-to-forget old categories are variable dynamically as consecutive learning tasks in the FCIL. It aggravates the difficulty of $\{\mathcal{V}_\kappa\}_{\kappa=1}^t$ to evaluate the heterogeneity of forgetting speeds for different old classes, due to the noisy probability predictions brought by easy-to-forget old classes. To this end, we enforce task-specific gradient means $\{\mathcal{V}_\kappa\}_{\kappa=1}^t$ to be sharper across different classes adaptively. $\{\mathcal{V}_\kappa\}_{\kappa=1}^t$ need to be more robust to address noisy predictions caused by easy-to-forget old classes when a series of tasks arrive continuously. Thus, Eq.~\eqref{eq: task_specific_gradient} is rewritten as: 
\begin{align}
	\mathcal{V}_{\kappa}^s = \frac{1}{\sum_{i=1}^B \mathbb{I}_{\mf{y}_{li}^t\in\mathcal{Y}_l^\kappa}} \sum\nolimits_{i=1}^B |\mathcal{V}_{li}^t|^{\frac{C_l^o}{C_l^o+C_l^t}}\cdot \mathbb{I}_{\mf{y}_{li}^t\in\mathcal{Y}_l^{\kappa}}. 
	\label{eq: task_specific_gradient_sharper}
\end{align}

The sharper task-specific gradient means $\{\mathcal{V}_\kappa^s\}_{\kappa=1}^t$ obtained via Eq.~\eqref{eq: task_specific_gradient_sharper} are then utilized to reweight standard cross-entropy loss $\mathcal{L}_{\mr{CE}}$ and we formulate the category-balanced gradient-adaptive compensation loss $\mathcal{L}_{\mr{CB}}$ as follows:
\begin{align}	
	\!\!\mathcal{L}_{\mr{CB}} \!=\! \frac{1}{B}\sum_{i=1}^B \frac{|\mathcal{V}_{li}^t|^{\frac{C_l^o}{C_l^o+C_l^t}}}{\sum_{\kappa=1}^t \mathcal{V}_\kappa^s\cdot\mathbb{I}_{\mf{y}_{li}^t\in\mathcal{Y}_l^{\kappa}}} \!\cdot\! \mathcal{D}_{\mr{CE}}(P_l^t(\mathbf{x}_{li}^t, \Theta_l^{r, t}), \mathbf{y}_{li}^t),\!\!
	\label{eq: gradient_compensation}
\end{align}
Obviously, Eq.~\eqref{eq: gradient_compensation} enforces the local model $\Theta_{l}^{r, t}$ to perform category-balanced gradient optimization adaptively, while addresses noisy predictions on easy-to-forget old categories via the sharper task-specific gradient means $\{\mathcal{V}_\kappa^s\}_{\kappa=1}^t$.

\textbf{$\bullet$ Category Gradient-Induced Semantic Distillation: }
After we initialize the local model $\Theta_l^{r, t}$ of $\mathcal{S}_l$ via the current global model $\Theta^{r, t}$ at the $r$-th global round, the probability predicted via $\Theta_l^{r, t}$ reflects semantic affinity among different categories during local training. In the FCIL, such semantic similarity relations between old and new categories are essential to address local catastrophic forgetting on hard-to-forget and easy-to-forget old classes. Thus, we propose a category gradient-induced semantic distillation loss $\mathcal{L}_{\mr{SD}}$ to explore category-semantic relation consistency across incremental tasks via considering category-balanced gradient propagation among new classes, hard-to-forget and easy-to-forget old classes. As shown in Fig.~\ref{fig: overview_of_our_model}, we feed a mini-batch $\{\mf{x}_{li}^t, \mf{y}_{li}^t\}_{i=1}^B\subset \mathcal{T}_l^t\cup\mathcal{M}_l$ into old model $\Theta_l^{t-1}$ and current local model $\Theta_l^{r, t}$, to predict the probabilities $P_l^{t-1}(\mathbf{x}_{li}^t, \Theta_l^{t-1})\in\mathbb{R}^{C^o}$ of old categories and $P_l^t(\mathbf{x}_{li}^t, \Theta_l^{r, t})\in\mathbb{R}^{C^o+C^t}$ of old and new categories. Obviously, the probabilities predicted by $\Theta_l^{t-1}$ and $\Theta_l^{r, t}$ indicate underlying semantic consistency of inter-class relations across different tasks. The proposed $\mathcal{L}_{\mr{SD}}$ can distill such intrinsic semantic consistency from old model $\Theta_l^{t-1}$ learned at the $(t\!-\!1)$-th task to current local model $\Theta_l^{r, t}$, and reweight the relation distillation of different tasks for category-balanced gradient propagation.

In order to distill semantic knowledge of old categories from $\Theta_l^{t-1}$ into $\Theta_l^{r, t}$, existing distillation methods \cite{Ahn_2021_ICCV, Hu_2021_CVPR, deng2021unbiased} only consider semantic consistency of old categories among $\Theta_l^{t-1}$ and $\Theta_l^{r, t}$. However, they neglect the intrinsic relation consistency between old and new categories. Besides, such semantic consistency of old categories can be significantly affected by the heterogeneous forgetting speeds of easy-to-forget and hard-to-forget old categories from different incremental tasks. To this end, the proposed loss $\mathcal{L}_{\mr{SD}}$ considers inter-class relations consistency within each incremental task, and reweights it via task-specific gradient means $\{\mathcal{V}_\kappa^s\}_{\kappa=1}^t$ to tackle heterogeneous forgetting speeds of old classes. To be specific, we utilize $P_l^{t-1}(\mathbf{x}_{li}^t, \Theta_l^{t-1})\in\mathbb{R}^{C^o}$ predicted by $\Theta_l^{t-1}$ to replace the first $C^o$ dimensions of $\mathbf{y}_{li}^t\in\mathbb{R}^{C^o+C^t}$, and obtain a variant $\mathbf{Y}_{li}^t\in\mathbb{R}^{C^o+C^t}$ to reflect inter-class semantic affinity between old and new classes via soft ground-truth labels. $\mathbf{Y}_{li}^t$ is then used to train local model $\Theta_l^{r, t}$ via $\mathcal{L}_{\mr{SD}}$:
\begin{align}
	\mathcal{L}_{\mr{SD}} & =\frac{1}{B}\sum_{i=1}^B \frac{|\mathcal{V}_{li}^t|^{\frac{C_l^o}{C_l^o+C_l^t}}}{\sum_{\kappa=1}^t \mathcal{V}_\kappa^s\cdot\mathbb{I}_{\mf{y}_{li}^t\in\mathcal{Y}_l^{\kappa}}} \cdot \big(\sum_{\kappa=1}^t \nonumber \\  
	&\mathcal{D}_{\mr{KL}}(P_l^t(\mathbf{x}_{li}^t, \Theta_l^{r, t})_{[C^{\kappa-1}+1:C^{\kappa}]}||(\mathbf{Y}_{li}^t)_{[C^{\kappa-1}+1:C^{\kappa}]})\big),
	\label{eq: relation_distillation}
\end{align}
where $\mathcal{D}_{\mr{KL}}(\mathbf{p}||\mathbf{q}) = \sum_i \mathbf{p}_i\log\frac{\mathbf{p}_i}{\mathbf{q}_i}$ is the Kullback-Leibler divergence. $P_l^t(\mathbf{x}_{li}^t, \Theta_l^{r, t})_{[C^{\kappa-1}+1:C^{\kappa}]}$ denotes probabilities from the $(C^{\kappa-1}\!+\!1)$-th to $C^{\kappa}$-th element in $P_l^t(\mathbf{x}_{li}^t, \Theta_l^{r, t})$, and its subscript $[C^{\kappa-1}\!+\!1:C^{\kappa}]$ is the class index learned at the $\kappa$-th incremental task. $(\mathbf{Y}_{li}^t)_{[C^{\kappa-1}+1:C^{\kappa}]}$ shares similar definition with $P_l^t(\mathbf{x}_{li}^t, \Theta_l^{r, t})_{[C^{\kappa-1}+1:C^{\kappa}]}$. Therefore, for the $l$-th local client $\mathcal{S}_l\in\mathcal{Z}_n$, the overall objective is expressed as: 
\begin{align}
	\mathcal{L}_{\mr{local}} = \gamma_1\mathcal{L}_{\mr{CB}} + \gamma_2\mathcal{L}_{\mr{SD}},
	\label{eq: overall_objective_local}
\end{align}
where $\gamma_1$ and $\gamma_2$ denote hyper-parameters. The local model $\Theta_l^{r, t}$ is trained via minimizing $\mathcal{L}_{\mr{local}}$ in Eq.~\eqref{eq: overall_objective_local} at the $r$-th global round. Then the global server $S_g$ collects all local models to aggregate them as the global model $\Theta^{r+1, t}$ for the next global round. When $t=1$, $\mathcal{L}_{\mr{SD}}$ lacks old model $\Theta_l^{t-1}$ to perform semantic distillation, thus we set $\gamma_1=1.0, \gamma_2 = 0$; otherwise, $\gamma_1=1.0, \gamma_2 = 1.0$. Moreover, the local clients belonging to $\mathcal{Z}_o$ and $\mathcal{Z}_a$ share the same optimization loss $\mathcal{L}_{\mr{local}}$ (\emph{i.e.}, Eq.~\eqref{eq: overall_objective_local}) with the local clients from $\mathcal{Z}_n$.

$\bullet$ \textbf{Task Transition Detection:}
In the FCIL settings, local clients are required to automatically identify when new classes arrive without prior knowledge. Then they can store old model $\Theta_l^{t-1}$ and update their exemplar memory $\mathcal{M}_l$ to train local model $\Theta_l^{r, t}$ via optimizing $\mathcal{L}_{\mr{local}}$ in Eq.~\eqref{eq: overall_objective_local}. To this end, an intuitive strategy is to distinguish whether the categories of new training data have already been learned by the current local client. Unfortunately, it cannot identify whether the labels of new training data belong to the new categories or old categories seen by other local clients, due to the Non-IID class distribution across local clients in the FCIL. Besides, the sharp performance decrease is considered as an indicator for local clients to receive new classes. However, this strategy neglects the fact that Non-IID class distribution across local clients brought by random determination of $\{\mathcal{Z}_o, \mathcal{Z}_n, \mathcal{Z}_a\}$ can result in significant performance degradation, even though local clients have not collected new classes.

To address these issues, we develop a task transition detection strategy for local clients to automatically distinguish when to collect new classes. Given training data $\mathcal{T}_l^t$ of the $t$-th task, each local client uses global model $\Theta^{r, t}$ (obtained from $\mathcal{S}_g$ at the $r$-th round) to compute average entropy $\mathcal{H}_l^{r, t}$:
\begin{equation}	
	\mathcal{H}_l^{r, t} = \frac{1}{N_l^t}\sum_{i=1}^{N_l^t}\mathcal{I}(P_l^t(\mf{x}_{li}^t, \Theta^{r, t})),
	\label{eq: task_transition}
\end{equation}
where $\mathcal{I}(\mathbf{p})=\sum_i \mathbf{p}_i\log \mathbf{p}_i$ denotes the entropy function. We argue that new categories are collected by local clients when there is a sudden rise for $\mathcal{H}_l^{r, t}$ (\emph{i.e.}, $\mathcal{H}_l^{r, t}-\mathcal{H}_l^{r-1, t}\geq r_e$). $r_e$ is set as $1.2$ empirically in this paper. After identifying new categories, local client will store old model $\Theta_l^{t-1}$, and update the exemplar memory $\mathcal{M}_l$ and $t$ via $t\leftarrow t + 1$.

\subsection{Global Anti-Forgetting}\label{sec: global_anti_forgetting}
As aforementioned, class imbalance between old and new categories at local side causes local catastrophic forgetting on old categories, which can be effectively tackled via Eq.~\eqref{eq: overall_objective_local}. However, the heterogeneous local forgetting brought by Non-IID class imbalance among different local clients enforces the global model to suffer from global catastrophic forgetting on old categories. Such global forgetting will exacerbate local forgetting on old categories to a certain extent. As a result, we consider tackling heterogeneous local forgetting across different clients from a global perspective \cite{Dong_2022_CVPR}. As introduced in Section~\ref{sec: local_anti_forgetting}, the proposed loss $\mathcal{L}_{\mr{SD}}$ in Eq.~\eqref{eq: overall_objective_local} can distill inter-class semantic relations within each incremental task from the old classification model $\Theta_l^{t-1}$ to current local model $\Theta_l^{r, t}$. Obviously, the relation distillation gain can be significantly improved via selecting a better old model $\Theta_l^{t-1}$ from a global perspective. Therefore, under local privacy preservation, the selection of best old model $\Theta_l^{t-1}$ from a global perspective is important to tackle heterogeneous local forgetting and achieve global anti-forgetting in the FCIL settings.

To this end, a trivial strategy is that each local client stores its own all old models when training local model in the $t$-th learning task. However, it only considers local data distribution to select $\Theta_l^{t-1}$ rather than a global perspective. Besides, it cannot ensure the selected $\Theta_l^{t-1}$ has the best performance to tackle global forgetting on all old categories, due to the Non-IID class distribution across different local clients (\emph{i.e.}, the label space $\mathcal{Y}_l^t\subset \mathcal{Y}^t$ of local client $\mathcal{S}_l$ is only a subset of $\mathcal{Y}^t$ in the $t$-th task). Thus, as shown in Fig.~\ref{fig: overview_of_our_model}, a proxy server $\mathcal{S}_p$ is developed to choose the best old model $\Theta^{t-1}$ for all local clients from a global perspective, while it considers privacy preservation of local clients.

To be specific, at the start of the $t$-th incremental task, local clients first employ task transition detection strategy (proposed in Section~\ref{sec: local_anti_forgetting}) to identify whether new categories are received. 
Considering privacy preservation at local side, we propose a prototype gradient communication mechanism to transmit perturbed prototype gradients of new categories from local clients to proxy server $\mathcal{S}_p$, after local clients have identified new categories. $\mathcal{S}_p$ then uses these gradients collected from local clients to reconstruct perturbed prototype images of new categories, and augments them via a self-supervised prototype augmentation strategy. After that, it evaluates the performance of global model $\Theta^{r, t}$ (obtained from $\mathcal{S}_g$) on these reconstructed perturbed images with representation augmentation to select the best one. 
When the best global model $\Theta^{r, t}$ is distributed from $\mathcal{S}_p$ to local clients, they consider it as the best old model to learn the next incremental task ($t\!+\!1$) via optimizing $\mathcal{L}_{\mr{SD}}$ in Eq.~\eqref{eq: overall_objective_local}.

$\bullet$ \textbf{Prototype Gradient Communication:}
When the $l$-th client $\mathcal{S}_l\in\mathcal{Z}_n\cup \mathcal{Z}_a$ has identified new classes in the $t$-th task $\mathcal{T}_l^t$ via task transition detention, it selects only one representative prototype image $\mf{x}_{lc^*}^t$ ($c=C_l^o\!+\!1, \cdots, C_l^o\!+\!C_l^t$) for each new category in $\mathcal{T}_l^t$. Note that the representation of $\mf{x}_{lc^*}^t$ is closest to the average feature of all images from the $c$-th category. The selected $C_l^t$ prototype images and corresponding labels $\{\mf{x}_{lc^*}^t, \mf{y}_{lc^*}^t\}_{c={C_l^o+1}}^{C_l^o+C_l^t}$ are forwarded into a gradient encoding network $\Gamma=\{\mathcal{W}_i\}_{i=1}^K$ with total $K$ layers to obtain their gradients $\{\nabla\Gamma_{lc}\}_{c={C_l^o+1}}^{C_l^o+C_l^t}$, where $\mathcal{W}_i$ denotes the weights of $i$-th layer and we set $K=4$ to reduce communication overhead. $\nabla_{\mathcal{W}_i}\Gamma_{lc}= \nabla_{\mathcal{W}_i} \mathcal{D}_{\mr{CE}}(P_l^t(\mf{x}_{lc^*}^t, \Gamma), \mf{y}_{lc^*}^t)$ represents the $i$-th gradient element of $\nabla\Gamma_{lc}$, where $P_l^t(\mf{x}_{lc^*}^t, \Gamma)$ denotes the predicted probability of $\mf{x}_{lc^*}^t$. These prototype gradients $\{\nabla\Gamma_{lc}\}_{c={C_l^o+1}}^{C_l^o+C_l^t}$ are transmitted from local clients $\mathcal{S}_l$ to proxy server $\mathcal{S}_p$ for prototype image reconstruction.

$\mathcal{S}_p$ constructs a gradient pool $\Omega^t = \cup_l\{\nabla\Gamma_{lc}\}_{c={C_l^o+1}}^{C_l^o+C_l^t}$ via randomly shuffling all collected prototype gradients from selected local clients, and we denote the number of prototype gradients in $\Omega^t$ as $N_g^t$. This shuffling strategy prevents $\mathcal{S}_p$ from tracking some special gradient distributions of selected local clients. As introduced in \cite{zhao2021DC, wang2021addressing}, the ground-truth label $y_s^t$ (the one-hot encoding vector of $y_s^t$ is $\mf{y}_s^t$) of the $s$-th element $\Omega_s^t$ in $\Omega^t$ can be determined by the gradient symbol of the last layer in $\Omega_s^t$. After initializing a noisy image $\bar{\mf{x}}_s^t$ via a standard Gaussian $\mathcal{N}(0, 1)$, we feed $\{\bar{\mf{x}}_s^t, \Omega_s^t, \mf{y}_s^t\}_{s=1}^{N_g^t}$ into $\Gamma=\{\mathcal{W}_i\}_{i=1}^K$ to reconstruct representative prototype images of new categories. Note that $\Gamma=\{\mathcal{W}_i\}_{i=1}^K$ used in proxy server and local clients has the same network framework. Therefore, we formulate the prototype image reconstruction loss $\mathcal{L}_{\mr{IR}}$ and the update of noisy image $\bar{\mf{x}}_s^t$ as follows:
\begin{align}
	\!\mathcal{L}_{\mr{IR}} &= \sum_{i=1}^{K} \|\nabla_{\mathcal{W}_i} \mathcal{D}_{\mr{CE}}(P^t(\bar{\mf{x}}_s^t, \Gamma), \mf{y}_s^t) - \nabla_{\mathcal{W}_i}\Omega_s^t \|^2, \\ 
	\bar{\mf{x}}_s^t &\leftarrow \bar{\mf{x}}_s^t - \mu\nabla_{\bar{\mf{x}}_s^t}\mathcal{L}_{\mr{IR}},
	\label{eq: reconstruction_loss}
\end{align}
where $P^t(\bar{\mf{x}}_s^t, \Gamma)$ represents the softmax probability obtained by $\Gamma$. $\nabla_{\mathcal{W}_i}\Omega_s^t$ denotes the ground-truth gradient with respect to the $i$-th layer $\mathcal{W}_i$. $\mu$ is the update rate to optimize $\bar{\mf{x}}_s^t$.

$\bullet$ \textbf{Self-Supervised Prototype Augmentation: }
When the selected local clients have detected new categories, they only transmit prototype gradients to $\mathcal{S}_p$ at the first global round of the $t$-th incremental task.  $\mathcal{S}_p$ reconstructs $N_g^t$ pairs of representative prototype images and labels $\{\bar{\mf{x}}_s^t, \mf{y}_s^t\}_{s=1}^{N_g^t}$ via Eq.~\eqref{eq: reconstruction_loss}. As aforementioned, only one representative image of each new category from selected local client is transmitted to the proxy server $\mathcal{S}_p$, thus the number $N_g^t$ of reconstructed prototype images is far insufficient to select the best old global model accurately via performance evaluation.

To this end, we propose a self-supervised prototype augmentation mechanism in this paper. Given a reconstructed image $\{\bar{\mf{x}}_s^t, \mf{y}_s^t\}$, we augment it in the latent feature space to obtain an augmented feature $\bar{\mf{f}}_s^t\in\mathbb{R}^d$:
\begin{align}
	\label{eq: augmented_feature}
	& \bar{\mf{f}}_s^t = \mathcal{F}(\bar{\mf{x}}_s^t) + \omega^t\mathcal{N}(0, \sigma^2), \\ \nonumber
	& s.t. ~\|\omega^t\mathcal{N}(0, \sigma^2)\|^2 \leq \xi \sum\nolimits_i^d \sigma_i^2,
\end{align}
where $\mathcal{F}(\bar{\mf{x}}_s^t)\in\mathbb{R}^d$ is the latent feature of $\bar{\mf{x}}_s^t$ that is extracted via feature extractor of the global model $\Theta^{r, t}$ (obtained from global server $\mathcal{S}_g$), and $d$ is the feature dimension of $\mathcal{F}(\bar{\mf{x}}_s^t)$. $\mathcal{N}(0, \sigma^2)$ is the Gaussian noise, $\sigma^2\in\mathbb{R}^d$ denotes the feature variance of all images labeled as $\mf{y}_s^t$, and $\sigma_i^2$ is the $i$-th element of $\sigma^2$. $\xi=0.1$ represents the radius of augmented feature space belonging to $\mf{y}_s^t$. $\omega^t$ adaptively determines the scale of feature augmentation, which is defined as follows: 
\begin{align}
	\omega^t = \frac{1}{C^o+C^t}(C^o\omega^{t-1} + \sum_{c=C^o+1}^{C^o+C^t}\frac{\mr{Tr}(\Delta_c^t)}{d}),
	\label{eq: augmentation_scale}
\end{align}
where $\Delta_c^t\in\mathbb{R}^{d\times d}$ is covariance matrix of features belonging to the $c$-th class in the $t$-th task. $\mr{Tr}(\cdot)$ is the trace function.

$\bullet$ \textbf{Selection of The Best Old Model: }
Given the prototype images $\{\bar{\mf{x}}_s^t, \mf{y}_s^t\}_{s=1}^{N_g^t}$ reconstructed via Eq.~\eqref{eq: reconstruction_loss}, the proxy server $\mathcal{S}_p$ utilizes self-supervised prototype augmentation to generate augmented prototype features $\{\bar{\mf{f}}_s^t, \mf{y}_s^t\}_{s=1}^{N_f^t}$ via Eq.~\eqref{eq: augmented_feature}, where $N_f^t = \upsilon N_g^t$ is the number of augmented features, and $\upsilon=5$ denotes the number of augmentation per image. When receiving the global model $\Theta^{r, t}$ ($r=1, \cdots, R$) from $\mathcal{S}_g$ in the $t$-th task, the proxy server $\mathcal{S}_p$ feeds all augmented prototype features $\{\bar{\mf{f}}_s^t, \mf{y}_s^t\}_{s=1}^{N_f^t}$ into the classifier of $\Theta^{r, t}$ to accurately monitor which global model has the best performance until the next task. In the $t$-th ($t\geq2$) incremental task, the best models of the $(t\!-\!1)$-th and $t$-th tasks (\emph{i.e.}, $\Theta^{t-1}$ and $\Theta^{t}$) are distributed from $\mathcal{S}_p$ to selected local clients at each global round. If $\mathcal{S}_l$ identifies new categories from the received training data $\mathcal{T}_l^{t+1}$ at the $t$-th task, $\Theta^{t}$ is considered as the best old model $\Theta_l^{t-1}$; otherwise, $\mathcal{S}_l$ regards $\Theta^{t-1}$ as $\Theta_l^{t-1}$ to optimize $\mathcal{L}_{\mr{SD}}$ in Eq.~\eqref{eq: overall_objective_local}.

\begin{algorithm}[t]
	\small
	\caption{Training Procedure of Our LGA Model.}
	\label{alg: training_pipline}
	\textbf{Initialize:} In the $t$-th ($t\geq2$) task, $\mathcal{S}_g$ randomly selects $n$ local clients $\{\mathcal{S}_{l_1}, ..., \mathcal{S}_{l_n}\}$ at the $r$-th global round. The selected clients have their private exemplar memories $\{\mathcal{M}_{l_1}, \cdots, \mathcal{M}_{l_n}\}$ and learning tasks $\{\mathcal{T}_{l_1}^t, \cdots, \mathcal{T}_{l_n}^t\}$.
	
	\textcolor{blue}{\textbf{All Local Clients:}} \\
	\For{$\mathcal{S}_l$ in $\{\mathcal{S}_1, \mathcal{S}_2, \cdots, \mathcal{S}_L\}$}{
		Compute average entropy $\mathcal{H}_l^{r, t}$ of $\mathcal{T}_l^t$ via Eq.~\eqref{eq: task_transition}; \\ 
		Update $\mathcal{M}_{l}$ via following iCaRL~\cite{Rebuffi_2017_CVPR}; \\
	}
	
	\textcolor{blue}{\textbf{Selected Clients (Local Anti-Forgetting in Section~\ref{sec: local_anti_forgetting}):}} \\
	Obtain $\Theta^{r, t}$ from global server $\mathcal{S}_g$ as local model;\\
	Obtain $\Theta^{t-1}$ and $\Theta^{t}$ from proxy server $\mathcal{S}_p$\;
	\For{$\mathcal{S}_{l}$ in $\{\mathcal{S}_{l_1}, \mathcal{S}_{l_2}, \cdots, \mathcal{S}_{l_n}\}$}{
		\emph{Task} = False\;
		\If{$\mathcal{H}_{l}^{r, t} - \mathcal{H}_{l}^{r-1, t} \geq r_e$}{
			\emph{Task} = True\;
		}
		\If{Task = \emph{True}}{
			$\Theta_{l}^{t-1} = \Theta^{t}$\;
		}
		\Else{
			$\Theta_{l}^{t-1} = \Theta^{t-1}$\;
		}
		\For{$\{\mathbf{X}_{lb}^t, \mathbf{Y}_{lb}^t\}\subset\mathcal{T}_{l}^t \cup \mathcal{M}_{l}$}{
			Optimize local model $\Theta_{l}^{r, t}$ via Eq.~\eqref{eq: overall_objective_local}; \\
		}
		\If{Task = \emph{True}}{
			$\nabla \Gamma_{l}^t = \{\}$\;
			\For{$c\in[C_{l}^o+1, \cdots, C_{l}^o+C_{l}^t]$}{
				Get perturbed image $(\mathbf{x}_{lc^*}^t, \mathbf{y}_{lc^*}^t)$ via Eq.~\eqref{eq: perturbation_reconstruction}; \\
				Obtain representative prototype gradient $\nabla \Gamma_{lc} = \cup_{\mathcal{W}_i} \nabla_{\mathcal{W}_i} \mathcal{D}_{\mr{CE}}(P_{l}^t(\mathbf{x}_{lc^*}^t, \Gamma), \mathbf{y}_{lc^*}^t)$; \\
				$\nabla \Gamma_{l}^t \leftarrow \nabla \Gamma_{l}^t \cup \nabla \Gamma_{lc}$; \\
			}
			Transmit these prototype gradients $\nabla \Gamma_{l}^t$ to $\mathcal{S}_p$; \\
		}
	}
	\textcolor{blue}{\textbf{Proxy Server (Global Anti-Forgetting in Section~\ref{sec: global_anti_forgetting}):}} \\
	Construct a shuffling pool $\Omega^t$ with $N_g^t$ gradients via collecting $\{\nabla \Gamma_{l_1}^t, \nabla \Gamma_{l_2}^t, ..., \nabla \Gamma_{l_n}^t\}$ from local clients; \\
	Obtain $\Theta^{r, t}$ from global server $\mathcal{S}_g$ as local model;\\
	\If{$\Omega^t\neq\emptyset$}{
		$\{\bar{\mathbf{X}}_p^t, \mathbf{Y}_p^t\} = \{\}$\;
		\For{$s = 1, \cdots, N_g^t$}{
			Reconstruct $(\bar{\mathbf{x}}_s^t, \mathbf{y}_s^t)$ via minimizing Eq.~\eqref{eq: reconstruction_loss}; \\
			$\{\bar{\mathbf{X}}_p^t, \mathbf{Y}_p^t\} \leftarrow \{\bar{\mathbf{X}}_p^t, \mathbf{Y}_p^t\} \cup (\bar{\mathbf{x}}_s^t, \mathbf{y}_s^t)$\;
		}
	}
	Augment prototype features of $\{\bar{\mathbf{X}}_p^t, \mathbf{Y}_p^t\}$ via Eq.~\eqref{eq: augmented_feature}; \\
	Select the best $\Theta^t$ via the results of $\Theta^{r, t}$ on $\{\bar{\mf{f}}_s^t, \mf{y}_s^t\}_{s=1}^{N_f^t}$; \\
	Send $\Theta^{t-1}$ and $\Theta^t$ to selected clients for local training. 
\end{algorithm}

$\bullet$ \textbf{Perturbed Prototype Image Construction: }
At the $t$-th task, the gradient encoding network $\Gamma$ and prototype gradients that are privately accessible to $\mathcal{S}_l$ and $\mathcal{S}_p$, may be stolen by malicious attackers to reconstruct raw prototype image $\{\mf{x}_{lc^*}^t, \mf{y}_{lc^*}^t\}\in\mathcal{T}_l^t$. Considering privacy protection, we introduce noisy perturbation to the latent feature space of prototype images. If attackers reconstruct perturbed prototype images, they can only capture useless information from them. To this end, we forward a prototype image $\{\mf{x}_{lc^*}^t, \mf{y}_{lc^*}^t\}\in\mathcal{T}_l^t$ to local model $\Theta_l^{r, t}$ that is well optimized via Eq.~\eqref{eq: overall_objective_local}, and update $\mf{x}_{lc^*}^t$ to get perturbed prototype image via introducing a Gaussian noise into its latent feature space. The perturbed prototype image $\mf{x}_{lc^*}^t$ is obtained by: 
\begin{align}
	\mathcal{L}_{\mr{PC}} &= \mathcal{D}_{\mr{CE}}(P_l^t(\mathcal{F}(\mf{x}_{lc^*}^t) + \epsilon\mathcal{N}(0, \sigma^2), \Theta_l^{r, t}), \mf{y}_{lc^*}^t), \\
	\mf{x}_{lc^*}^t &\leftarrow \mf{x}_{lc^*}^t - \mu \nabla_{\mf{x}_{lc^*}^t}\mathcal{L}_{\mr{PC}},
	\label{eq: perturbation_reconstruction}
\end{align}
where $\mathcal{F}(\mf{x}_{lc^*}^t)$ is the latent feature of $\mf{x}_{lc^*}^t$. $P_l^t(\mathcal{F}(\mf{x}_{lc^*}^t) + \epsilon\mathcal{N}(0, \sigma^2), \Theta_l^{r, t})$ denotes the probability of latent feature $\mathcal{F}(\mf{x}_{lc^*}^t)$ with Gaussian noise $\mathcal{N}(0, \sigma^2)$, which is predicted by local model $\Theta_l^{r, t}$. $\sigma^2$ is the feature variance of all images labeled as $\mf{y}_{lc^*}^t$.
$\epsilon=0.1$ is the parameter to balance the effect of Gaussian noise. $\mu$ is the learning rate to update $\mf{x}_{lc^*}^t$.

\textbf{Security Assumptions:}
We follow some mild assumptions that the proxy server is hosted by a trusted third party and is also responsible for initializing an encoding model $\Gamma$ before training FL framework. Our mechanism will not communicate $\Gamma$ during FL training, while each selected client uses its own channel to communicate gradients with the proxy server. There is no communication channel between clients, and the proxy server is required not to share received gradients to the global server or other parties. Moreover, all channels are reliable and protected by crypto protocols. With these assumptions, we believe that the proxy server is secure and hard to be attacked. Even if malicious attackers could compromise several selected clients and get $\Gamma$, they cannot steal the gradients of other clients since the communication channels are separated. In the worst case, attackers could eavesdrop gradients in some ways, but they can only recover perturbed prototype exemplars (see Fig.~\ref{fig: prototype_samples}), where the privacy of local clients is protected via our proposed mechanism.

\subsection{Training Pipeline}
As presented in Algorithm~\ref{alg: training_pipline}, we introduce the overall training pipeline of proposed LGA model. Specifically, when receiving training data of the first incremental task, all local clients update their private exemplar memory $\mathcal{M}_l$ via following iCaRL \cite{Rebuffi_2017_CVPR}, and use their private data to compute the average entropy $\mathcal{H}_l^{r, t}$ via Eq.~\eqref{eq: task_transition}. In the FCIL, some local clients randomly selected by global server $\mathcal{S}_g$ are required to perform local training at each global round. If these selected local clients have detected new categories via the task transition detection, they will generate perturbed prototype images belonging to new categories via optimizing Eq.~\eqref{eq: perturbation_reconstruction}, and transmit these prototype gradients to $\mathcal{S}_p$ via prototype gradient communication strategy. Then $\mathcal{S}_p$ uses these prototype gradients received from local clients to reconstruct perturbed prototype images via Eq.~\eqref{eq: reconstruction_loss}. The reconstructed images are augmented by $\mathcal{S}_p$ in the feature space, which are then employed to pick the best global model $\Theta^t$ until the next task. In the $t$-th ($t\geq2$) task, $\mathcal{S}_p$ distributes the best global models of the $(t\!-\!1)$-th and $t$-th tasks (\emph{i.e.}, $\Theta^{t-1}$ and $\Theta^{t}$) to local client $\mathcal{S}_l$. If $\mathcal{S}_l$ detects new categories via task transition detection, it regards $\Theta^{t}$ as the best old model $\Theta_l^{t-1}$ to perform local training via minimizing Eq.~\eqref{eq: overall_objective_local}; otherwise, $\Theta^{t-1}$ is used as $\Theta_l^{t-1}$ to train local model $\Theta_l^{r, t}$. After that, all updated local models $\Theta_l^{r, t}$ are collected by $\mathcal{S}_g$ to aggregate them as global model $\Theta^{r+1, t}$ for the next round.

\begin{table*}[t]
\centering
\setlength{\tabcolsep}{1.72mm}
\caption{Comparisons in terms of accuracy on CIFAR-100 \cite{krizhevsky2009learning} and MiniImageNet \cite{NIPS2016_90e13578} datasets when setting the number of continual tasks as $T=5$. }  
\scalebox{0.99}{
\begin{tabular}{c|ccccc|>{\columncolor{lightgray}}c|>{\columncolor{lightgray}}l|ccccc|>{\columncolor{lightgray}}c|>{\columncolor{lightgray}}l}
	\toprule
	\multirow{2}{*}{Methods} & \multicolumn{7}{c|}{CIFAR-100 \cite{krizhevsky2009learning}} & \multicolumn{7}{c}{MiniImageNet \cite{NIPS2016_90e13578}}  \\
	& 20 & 40 & 60 & 80 & 100 & Avg. (\%) & \makecell[c]{Imp. (\%)} & 20 & 40 & 60 & 80 & 100 & Avg. (\%) & \makecell[c]{Imp. (\%)}  \\
	\midrule
	iCaRL \cite{Rebuffi_2017_CVPR} + FL & 77.0 & 59.6 & 51.9 & 44.4 & 39.6 & 54.5 & $\Uparrow$ 18.5 & 73.5& 56.2& 46.2& 40.2& 35.5 & 50.3 & $\Uparrow$ 18.4  \\
	BiC \cite{wu2019large} + FL & 78.4 & 60.4 & 53.2 & 47.5& 41.2 & 56.1 & $\Uparrow$ 16.9 & 72.6& 56.8& 49.2& 43.5& 38.7 & 52.2 & $\Uparrow$ 16.5  \\
	PODNet \cite{10.1007/978-3-030-58565-5_6} + FL &77.6& 62.1& 56.3& 50.8& 43.3 & 58.0 & $\Uparrow$ 15.0 & 73.1& 58.4& 53.2& 46.5& 43.4 & 54.9 & $\Uparrow$ 13.8  \\
	DDE \cite{Hu_2021_CVPR} + iCaRL \cite{Rebuffi_2017_CVPR} + FL &77.0& 60.2& 55.7& 49.3& 42.5 & 56.9 & $\Uparrow$ 16.1 & 72.3& 57.2& 51.7& 44.3& 41.3& 53.4 & $\Uparrow$ 15.3  \\
	GeoDL \cite{Christian2021MGeoCont} + iCaRL \cite{Rebuffi_2017_CVPR} + FL & 72.5& 61.1& 54.0& 49.5& 44.5& 56.3 & $\Uparrow$ 16.7 & 71.8& 59.6& 52.3& 46.1 & 42.5 & 54.5 & $\Uparrow$ 14.2  \\
	SS-IL \cite{Ahn_2021_ICCV} + FL & 78.1& 61.8& 52.8& 48.8& 46.0 & 57.5 & $\Uparrow$ 15.5 & 66.5& 52.1& 42.6& 36.7& 36.5 & 46.9 & $\Uparrow$ 21.8 \\
	
	DyTox \cite{Douillard_2022_CVPR} + FL & 78.8& 70.5& 63.9& 59.9& 55.9 & 65.8 & $\Uparrow$ 7.2 & 69.6& 64.2 & 59.1 & 53.4& 48.5& 59.0 & $\Uparrow$ 9.7  \\ 
	AFC \cite{Kang_2022_CVPR} + FL & 71.1& 63.8& 58.4& 53.6& 46.4 & 58.3 & $\Uparrow$ 14.7 & 78.0& 64.5& 57.0& 51.3& 47.3 & 59.6 & $\Uparrow$ 9.1  \\
	GLFC \cite{Dong_2022_CVPR} & \textcolor[rgb]{0.698,0.133,0.133}{\textbf{83.7}}& 75.5& 66.5& 62.1& 53.8 & 68.3 & $\Uparrow$ 4.7 & 
	\textcolor[rgb]{0.698,0.133,0.133}{\textbf{79.7}}& 73.4& 65.2& 58.1& 51.8 & 65.6 & $\Uparrow$ 3.1  \\
	
	\midrule
	\midrule
	Ours-w/oCBL & 83.3& 72.6& 65.6& 58.2& 53.8 & 66.7 & $\Uparrow$ 5.2 & 79.0& 71.2& 61.1& 54.6& 50.0 & 63.2 & $\Uparrow$ 5.5  \\
	Ours-w/oSDL & 83.2& 66.7& 59.0& 49.5& 44.8 & 60.4 & $\Uparrow$ 12.6 &78.4& 65.1& 60.6& 53.7& 49.3& 61.4 & $\Uparrow$ 7.3  \\
	Ours-w/oPSR & 82.7& 76.7& 71.5& 66.4& 62.3& 71.9 & $\Uparrow$ 1.1 &78.5& 70.3& 67.6& 61.2& 61.5 & 67.8 & $\Uparrow$ 0.9  \\
	\textbf{Ours (LGA)} & 83.3& \textcolor[rgb]{0.698,0.133,0.133}{\textbf{77.3}}& \textcolor[rgb]{0.698,0.133,0.133}{\textbf{72.8}}& 
	\textcolor[rgb]{0.698,0.133,0.133}{\textbf{67.8}}& 
	\textcolor[rgb]{0.698,0.133,0.133}{\textbf{63.7}} & \textcolor[rgb]{0.698,0.133,0.133}{\textbf{73.0}} & --- & 
	78.9 & 
	\textcolor[rgb]{0.698,0.133,0.133}{\textbf{75.5}}& 
	\textcolor[rgb]{0.698,0.133,0.133}{\textbf{68.1}}& 
	\textcolor[rgb]{0.698,0.133,0.133}{\textbf{62.1}} & 
	\textcolor[rgb]{0.698,0.133,0.133}{\textbf{61.9}} & \textcolor[rgb]{0.698,0.133,0.133}{\textbf{68.7}} & --- \\ 
	
	\bottomrule
\end{tabular}}
\label{tab: exp_all_datasets_task5}
\end{table*}

\begin{table*}[t]
\centering
\setlength{\tabcolsep}{1.7mm}
\caption{Comparisons in terms of accuracy on TinyImageNet \cite{Tiny_imagenet} and ImageNet-1000 \cite{5206848} datasets when setting the number of learning tasks as $T=5$. } 
\scalebox{0.99}{
\begin{tabular}{c|ccccc|>{\columncolor{lightgray}}c|>{\columncolor{lightgray}}l|ccccc|>{\columncolor{lightgray}}c|>{\columncolor{lightgray}}l}
	\toprule
	\multirow{2}{*}{Methods} & \multicolumn{7}{c|}{TinyImageNet \cite{Tiny_imagenet}} & \multicolumn{7}{c}{ImageNet-1000 \cite{5206848} } \\
	& 40 & 80 & 120 & 160 & 200 & Avg. (\%) & \makecell[c]{Imp. (\%)} & 200 & 400 & 600 & 800 & 1000 & Avg. (\%) & \makecell[c]{Imp. (\%)}  \\
	\midrule
	iCaRL \cite{Rebuffi_2017_CVPR} + FL & 65.0& 48.0& 42.7& 38.7& 35.0 & 45.9 & $\Uparrow$ 8.6 & 70.1& 59.3& 54.8& 50.7& 40.4& 55.1 & $\Uparrow$ 8.4 \\
	BiC \cite{wu2019large} + FL & 65.7& 48.7& 43.0& 40.3& 35.7& 46.7 & $\Uparrow$ 7.8 & 68.2 & 57.2 & 49.6 & 40.3 & 37.5 & 50.6 & $\Uparrow$ 12.9\\
	PODNet \cite{10.1007/978-3-030-58565-5_6} + FL & 66.0 & 50.3& 44.7& 41.3& 37.0& 47.9 & $\Uparrow$ 6.6 & 68.5 & 58.1 & 50.3 & 41.8 & 39.4 & 51.6 & $\Uparrow$ 11.9 \\
	DDE \cite{Hu_2021_CVPR} + iCaRL \cite{Rebuffi_2017_CVPR} + FL & 63.0& 51.3& 45.3& 41.0& 36.0 & 47.3 & $\Uparrow$ 7.2 & 67.3 & 52.3 & 47.1 & 40.2 & 34.8 & 48.3 & $\Uparrow$ 15.2 \\
	GeoDL \cite{Christian2021MGeoCont} + iCaRL \cite{Rebuffi_2017_CVPR} + FL & 65.3& 50.0& 45.0& 40.7 &36.0 & 47.4 & $\Uparrow$ 7.1 & 67.1 & 51.3 & 45.8 & 41.1 & 35.3 & 48.1 & $\Uparrow$ 15.4 \\
	SS-IL \cite{Ahn_2021_ICCV} + FL & 65.0 &42.3 &38.3& 35.0& 30.3& 42.2& $\Uparrow$ 12.3 & 66.4 & 47.3 & 40.7 & 35.8 & 31.2 & 44.3 & $\Uparrow$ 19.2 \\
	
	DyTox \cite{Douillard_2022_CVPR} + FL & 58.6& 43.1& 41.6 & 37.2 & 32.9 & 42.7 & $\Uparrow$ 11.8 & 64.7& 58.8& 51.6& 46.9& 43.7 & 53.1 & $\Uparrow$ 10.4 \\ 
	AFC \cite{Kang_2022_CVPR} + FL &62.5& 52.1& 45.7& 43.2& 35.7 & 47.8 & $\Uparrow$ 6.7 & 65.6 & 63.1 & 61.9& 55.7 & 47.4& 58.7 & $\Uparrow$ 4.8 \\
	GLFC \cite{Dong_2022_CVPR} & 66.0& 55.3& 49.0& 45.0& 40.3& 51.1& $\Uparrow$ 3.4 & 70.3 & 64.0 & 63.1 & 56.5 & 49.3 & 60.6 & $\Uparrow$ 2.9 \\
	
	\midrule
	\midrule
	Ours-w/oCBL & 66.1& 59.7& 47.9& 42.3& 37.3 & 50.7 & $\Uparrow$ 3.8 & 70.3 & 63.7 & 61.8 & 54.8 & 46.3 & 59.4 & $\Uparrow$ 4.1 \\
	Ours-w/oSDL &\textcolor[rgb]{0.698,0.133,0.133}{\textbf{71.2}}& 49.8& 40.7& 33.6& 30.7 & 45.2& $\Uparrow$ 9.3 & 70.5 & 59.2 & 55.1 & 48.7 & 43.4 & 55.4 & $\Uparrow$ 8.1 \\
	Ours-w/oPSR &67.5& 50.6& 43.6& 35.6& 31.4 & 45.7 & $\Uparrow$ 8.8 & 70.5 & 64.5 & 62.1 & 55.3 & 47.9& 60.1 & $\Uparrow$ 3.4 \\
	\textbf{Ours (LGA)} &
	67.7 & 
	\textcolor[rgb]{0.698,0.133,0.133}{\textbf{59.8}}& 
	\textcolor[rgb]{0.698,0.133,0.133}{\textbf{53.5}}& 
	\textcolor[rgb]{0.698,0.133,0.133}{\textbf{47.9}}& 
	\textcolor[rgb]{0.698,0.133,0.133}{\textbf{43.8}} & \textcolor[rgb]{0.698,0.133,0.133}{\textbf{54.5}} & --- & \textcolor[rgb]{0.698,0.133,0.133}{\textbf{70.6}} & \textcolor[rgb]{0.698,0.133,0.133}{\textbf{68.3}} & \textcolor[rgb]{0.698,0.133,0.133}{\textbf{66.9}} & \textcolor[rgb]{0.698,0.133,0.133}{\textbf{59.6}} & \textcolor[rgb]{0.698,0.133,0.133}{\textbf{52.1}} & \textcolor[rgb]{0.698,0.133,0.133}{\textbf{63.5}} & --- \\ 
	
	\bottomrule
\end{tabular}}
\label{tab: exp_imagenet_datasets_task5}
\end{table*}

\begin{table*}[t]
\centering
\setlength{\tabcolsep}{2.77mm}
\caption{Comparison experiments in terms of accuracy on CIFAR-100 dataset \cite{krizhevsky2009learning} when setting the number of consecutive learning tasks as $T=10$. } 
\scalebox{0.9999}{
	\begin{tabular}{c|cccccccccc|>{\columncolor{lightgray}}c|>{\columncolor{lightgray}}l}
		\toprule
		Methods & 10 & 20 & 30 & 40 & 50 & 60 & 70 & 80 & 90 & 100 & Avg. (\%) & \makecell[c]{Imp. (\%)} \\
		\midrule
		iCaRL \cite{Rebuffi_2017_CVPR} + FL & 89.0 & 55.0 & 57.0 & 52.3 & 50.3 & 49.3 & 46.3 & 41.7 & 40.3 & 36.7 & 51.8 & $\Uparrow$ 21.7 \\ 
		BiC \cite{wu2019large} + FL & 88.7 & 63.3 & 61.3 & 56.7 & 53.0 & 51.7 & 48.0 & 44.0 & 42.7 & 40.7 & 55.0 & $\Uparrow$ 18.5 \\ 
		PODNet \cite{10.1007/978-3-030-58565-5_6} + FL & 89.0 & 71.3 & 69.0 & 63.3 & 59.0 & 55.3 & 50.7 & 48.7 & 45.3 & 45.0 & 59.7 & $\Uparrow$ 13.8 \\
		DDE \cite{Hu_2021_CVPR} + iCaRL \cite{Rebuffi_2017_CVPR} + FL & 88.0 & 70.0 & 67.3 & 62.0 & 57.3 & 54.7 & 50.3 & 48.3 & 45.7 & 44.3 & 58.8 & $\Uparrow$ 14.7 \\
		GeoDL \cite{Christian2021MGeoCont} + iCaRL \cite{Rebuffi_2017_CVPR} + FL & 87.0 & 76.0 & 70.3 & 64.3 & 60.7 & 57.3 & 54.7 & 50.3 & 48.3 & 46.3 & 61.5 & $\Uparrow$ 12.0 \\
		SS-IL \cite{Ahn_2021_ICCV} + FL & 88.3 & 66.3 & 54.0 & 54.0 & 44.7 & 54.7 & 50.0 & 47.7 & 45.3 & 44.0 & 54.9 & $\Uparrow$ 18.6 \\
		
		DyTox \cite{Douillard_2022_CVPR} + FL & 86.2 & 76.9 & 73.3 & 69.5 & 62.1 & 62.7 & 58.1 & 57.2& 55.4& 52.1 & 65.4 & $\Uparrow$ 8.1 \\
		AFC \cite{Kang_2022_CVPR} + FL & 85.6 & 73.0 & 65.1 & 62.4 & 54.0 & 53.1 &51.9 & 47.0 &46.1 &43.6 & 58.2 & $\Uparrow$ 15.3 \\
		GLFC \cite{Dong_2022_CVPR} & \textcolor[rgb]{0.698,0.133,0.133}{\textbf{90.0}} & 82.3 & 77.0 & 72.3 & 65.0 & 66.3 & 59.7 & 56.3 & 50.3 & 50.0 & 66.9 & $\Uparrow$ 6.6 \\
		
		\midrule
		\midrule
		Ours-w/oCBL & 88.7& 81.7& 78.0& 72.8& 67.6& 64.3& 61.9& 56.7& 57.1& 54.8 & 68.4 & $\Uparrow$ 5.1 \\
		Ours-w/oSDL & 89.9& 62.2& 62.9& 61.1& 54.2& 51.1& 50.2& 45.8& 45.3& 41.7& 56.4 & $\Uparrow$ 17.1 \\
		Ours-w/oPSR & 88.5& 82.1& 79.0& 75.1& 70.7& 69.9& 63.3& 63.0& 60.8& 59.7 & 71.2 & $\Uparrow$ 2.3 \\
		\textbf{Ours (LGA)} &89.6& 
		\textcolor[rgb]{0.698,0.133,0.133}{\textbf{83.2}}& 
		\textcolor[rgb]{0.698,0.133,0.133}{\textbf{79.3}}& 
		\textcolor[rgb]{0.698,0.133,0.133}{\textbf{76.1}}& 
		\textcolor[rgb]{0.698,0.133,0.133}{\textbf{72.9}}& 
		\textcolor[rgb]{0.698,0.133,0.133}{\textbf{71.7}}& 
		\textcolor[rgb]{0.698,0.133,0.133}{\textbf{68.4}}& 
		\textcolor[rgb]{0.698,0.133,0.133}{\textbf{65.7}}& 
		\textcolor[rgb]{0.698,0.133,0.133}{\textbf{64.7}}& 
		\textcolor[rgb]{0.698,0.133,0.133}{\textbf{62.9}} & \textcolor[rgb]{0.698,0.133,0.133}{\textbf{73.5}} & --- \\ 
		
		\bottomrule
\end{tabular}}
\label{tab: exp_cifar100_task10}
\end{table*}

\begin{table*}[t]
	\centering
	\setlength{\tabcolsep}{2.77mm}
	\caption{Comparison experiments in terms of accuracy on MiniImageNet dataset \cite{NIPS2016_90e13578} when setting the number of consecutive learning tasks as $T=10$. } 
	\scalebox{0.9999}{
		\begin{tabular}{c|cccccccccc|>{\columncolor{lightgray}}c|>{\columncolor{lightgray}}l}
			\toprule
			Methods & 10 & 20 & 30 & 40 & 50 & 60 & 70 & 80 & 90 & 100 & Avg. (\%) & \makecell[c]{Imp. (\%)}  \\
			\midrule
			iCaRL \cite{Rebuffi_2017_CVPR} + FL & 74.0 & 62.3 & 56.3 & 47.7 & 46.0 & 40.3 & 37.7 & 34.3 & 33.3 & 32.7 & 46.5 & $\Uparrow$ 21.0 \\ 
			BiC \cite{wu2019large} + FL & 74.3 & 63.0 & 57.7 & 51.3 & 48.3 & 46.0 & 42.7 & 37.7 & 35.3 & 34.0 & 49.0 & $\Uparrow$ 18.5 \\ 
			PODNet \cite{10.1007/978-3-030-58565-5_6} + FL & 74.3 & 64.0 & 59.0 & 56.7 & 52.7 & 50.3 & 47.0 & 43.3 & 40.0 & 38.3 & 52.6 & $\Uparrow$ 14.9 \\ 
			DDE \cite{Hu_2021_CVPR} + iCaRL \cite{Rebuffi_2017_CVPR} + FL & 76.0 & 57.7 & 58.0 & 56.3 & 53.3 & 50.7 & 47.3 & 44.0 & 40.7 & 39.0 & 52.3 & $\Uparrow$ 15.2 \\
			GeoDL \cite{Christian2021MGeoCont} + iCaRL \cite{Rebuffi_2017_CVPR} + FL & 74.0 & 63.3 & 54.7 & 53.3 & 50.7 & 46.7 & 41.3 & 39.7 & 38.3 & 37.0 & 50.0 & $\Uparrow$ 17.5 \\
			SS-IL \cite{Ahn_2021_ICCV} + FL & 69.7 & 60.0 & 50.3 & 45.7 & 41.7 & 44.3 & 39.0 & 38.3 & 38.0 & 37.3 & 46.4 & $\Uparrow$ 21.1 \\
			
			DyTox \cite{Douillard_2022_CVPR} + FL & 76.3 & 68.3& 64.8& 58.6 & 45.4 & 41.3 & 39.7 & 37.1 & 36.2 & 35.3 & 50.3 & $\Uparrow$ 17.2 \\
			AFC \cite{Kang_2022_CVPR} + FL &82.5& 74.1& 66.8& 60.0& 48.0& 44.3& 42.5& 40.9& 39.0& 36.1 & 53.4 & $\Uparrow$ 14.1 \\
			
			GLFC \cite{Dong_2022_CVPR} & 73.0 & 69.3 & 68.0 & 61.0 & 58.3 & 54.0 & 51.3 & 48.0 & 44.3 & 42.7 & 57.0 & $\Uparrow$ 10.5 \\
			\midrule
			\midrule
			Ours-w/oCBL &82.5& 73.7& 70.6& 69.8& 62.6& 58.9& 54.4& 47.6& 46.6& 43.7 & 61.0 & $\Uparrow$ 6.5 \\
			Ours-w/oSDL &82.4& 59.6& 56.5& 54.6& 48.1& 45.8& 40.3& 39.9& 35.7& 29.2 & 49.2 & $\Uparrow$ 18.3 \\
			Ours-w/oPSR &82.7& 71.8& 68.9& 69.8& 65.0& 62.4& 59.0& 58.0& 55.8& 56.3 & 65.0 & $\Uparrow$ 2.5 \\
			\textbf{Ours (LGA)} & \textcolor[rgb]{0.698,0.133,0.133}{\textbf{83.0}}& \textcolor[rgb]{0.698,0.133,0.133}{\textbf{74.2}}& \textcolor[rgb]{0.698,0.133,0.133}{\textbf{72.3}}& \textcolor[rgb]{0.698,0.133,0.133}{\textbf{72.2}}& \textcolor[rgb]{0.698,0.133,0.133}{\textbf{68.1}}& \textcolor[rgb]{0.698,0.133,0.133}{\textbf{65.8}}& 
			\textcolor[rgb]{0.698,0.133,0.133}{\textbf{64.0}} & \textcolor[rgb]{0.698,0.133,0.133}{\textbf{59.6}}& \textcolor[rgb]{0.698,0.133,0.133}{\textbf{58.4}}& 
			\textcolor[rgb]{0.698,0.133,0.133}{\textbf{57.5}} & \textcolor[rgb]{0.698,0.133,0.133}{\textbf{67.5}} & --- \\ 
			\bottomrule
	\end{tabular} } 	
	\label{tab: exp_imageNet_subset_task10} 
\end{table*}

\begin{table*}[t]
	\centering
	\setlength{\tabcolsep}{2.77mm}
	\caption{Comparison experiments in terms of accuracy on TinyImageNet dataset \cite{Tiny_imagenet} when setting the number of consecutive learning tasks as $T=10$. }  
	\scalebox{0.9999}{
		\begin{tabular}{c|cccccccccc|>{\columncolor{lightgray}}c|>{\columncolor{lightgray}}l}
			\toprule
			Methods & 20 & 40 & 60 & 80 & 100 & 120 & 140 & 160 & 180 & 200 & Avg. (\%) & \makecell[c]{Imp. (\%)} \\
			\midrule
			iCaRL \cite{Rebuffi_2017_CVPR} + FL & 63.0 & 53.0 & 48.0 & 41.7 & 38.0 & 36.0 & 33.3 & 30.7 & 29.7 & 28.0 & 40.1 & $\Uparrow$ 13.1 \\ 
			BiC \cite{wu2019large} + FL & 65.3 & 52.7 & 49.3 & 46.0 & 40.3 & 38.3 & 35.7 & 33.0 & 31.7 & 29.0 & 42.1 & $\Uparrow$ 11.1 \\ 
			PODNet \cite{10.1007/978-3-030-58565-5_6} + FL & 66.7 & 53.3 & 50.0 & 47.3 & 43.7 & 42.7 & 40.0 & 37.3 & 33.7 & 31.3 & 44.6 & $\Uparrow$ 8.6 \\ 
			DDE \cite{Hu_2021_CVPR} + iCaRL \cite{Rebuffi_2017_CVPR} + FL & 69.0 & 52.0 & 50.7 & 47.0 & 43.3 & 42.0 & 39.3 & 37.0 & 33.0 & 31.3 & 44.5 & $\Uparrow$ 8.7 \\ 
			GeoDL \cite{Christian2021MGeoCont} + iCaRL \cite{Rebuffi_2017_CVPR} + FL & 66.3 & 54.3 & 52.0 & 48.7 & 45.0 & 42.0 & 39.3 & 36.0 & 32.7 & 30.0 & 44.6 & $\Uparrow$ 8.6 \\ 
			SS-IL \cite{Ahn_2021_ICCV} + FL & 62.0 & 48.7 & 40.0 & 38.0 & 37.0 & 35.0 & 32.3 & 30.3 & 28.7 & 27.0 & 37.9 & $\Uparrow$ 15.3 \\
			
			DyTox \cite{Douillard_2022_CVPR} + FL & 73.2& 66.6& 48.0& 47.1& 41.6 & 40.8& 37.4 & 36.2 & 32.8 & 30.6 & 45.4 & $\Uparrow$ 7.8 \\
			AFC \cite{Kang_2022_CVPR} + FL & 73.7& 59.1& 50.8& 43.1& 37.0& 35.2& 32.6& 32.0& 28.9& 27.1 & 42.0 & $\Uparrow$ 11.2 \\
			
			GLFC \cite{Dong_2022_CVPR} & 66.0 & 58.3 & 55.3 & 51.0 & 47.7 & 45.3 & 43.0 & 40.0 & 37.3 & 35.0 & 47.9 & $\Uparrow$ 5.3 \\ 
			\midrule
			\midrule
			Ours-w/oCBL &70.1& 60.0& 57.1& 54.1& 47.8& 47.2& 41.8& 36.9 & 33.4 & 32.8 & 48.1 & $\Uparrow$ 5.1 \\
			Ours-w/oSDL &\textcolor[rgb]{0.698,0.133,0.133}{\textbf{73.8}}& 47.0& 53.1& 51.1& 47.2& 42.4& 40.2& 35.1& 31.6& 31.0 & 45.3 & $\Uparrow$ 7.9 \\
			Ours-w/oPSR &72.5& 59.1& 55.6& 52.1& 47.9& 46.7& 38.6& 33.1& 28.4& 24.5& 45.9 & $\Uparrow$ 7.3 \\ 
			\textbf{Ours (LGA)} & 70.3& 
			\textcolor[rgb]{0.698,0.133,0.133}{\textbf{64.0}}& \textcolor[rgb]{0.698,0.133,0.133}{\textbf{60.3}}& \textcolor[rgb]{0.698,0.133,0.133}{\textbf{58.0}}& \textcolor[rgb]{0.698,0.133,0.133}{\textbf{55.8}}& \textcolor[rgb]{0.698,0.133,0.133}{\textbf{53.1}}& 
			\textcolor[rgb]{0.698,0.133,0.133}{\textbf{47.9}} & \textcolor[rgb]{0.698,0.133,0.133}{\textbf{45.3}} &
			\textcolor[rgb]{0.698,0.133,0.133}{\textbf{39.8}} & \textcolor[rgb]{0.698,0.133,0.133}{\textbf{37.3}} & \textcolor[rgb]{0.698,0.133,0.133}{\textbf{53.2}} & --- \\
			\bottomrule
	\end{tabular} } 	
	\label{tab: exp_tinyimageNet_task10}
\end{table*}

\section{Experiments}

\subsection{Datasets and Evaluation Metrics}
We introduce comparison experiments on CIFAR-100 \cite{krizhevsky2009learning}, MiniImageNet \cite{NIPS2016_90e13578}, TinyImageNet \cite{Tiny_imagenet} and ImageNet-1000 \cite{5206848} datasets to illustrate the effectiveness of our LGA model.

\textbf{CIFAR-100} \cite{krizhevsky2009learning} consists of 100 different categories with total 60,000 samples, where each category has 500 training samples and 100 test images, and the image size is $32\times 32$. 
\textbf{MiniImageNet} \cite{NIPS2016_90e13578} is composed of 100 categories that are generated from ImageNet \cite{5206848}. The images of each category are split into a training set with 500 images and a test set with 100 images.  
\textbf{TinyImageNet} \cite{Tiny_imagenet} consists of 100,000 samples from 200 categories. We assign 500 training images and 50 test images for each category, and downsize each image to $64\times 64$. 
\textbf{ImageNet-1000} \cite{5206848} is a large-scale dataset including $1,000$ categories. For each class, we use about $1,300$ images for training and 50 images for testing.

\textbf{Evaluation Metric:} In this work, we utilize three kinds of metrics (\emph{i.e.}, top-1 accuracy, F1 score and recall) to show the superior performance of our proposed LGA model.

\begin{table*}[t]
	\centering
	\setlength{\tabcolsep}{0.9mm}
	\caption{Comparison experiments in terms of accuracy on CIFAR-100 dataset \cite{krizhevsky2009learning} when setting the number of consecutive learning tasks as $T=20$. }
	\scalebox{0.905}{
		\begin{tabular}{c|cccccccccccccccccccc|>{\columncolor{lightgray}}c|>{\columncolor{lightgray}}l}
			\toprule
			Methods & 5 & 10 & 15 & 20 & 25 & 30 & 35 & 40 & 45 & 50 & 55 & 60 & 65 & 70 & 75 & 80 & 85 & 90 & 95 & 100 & Avg. (\%) & \makecell[c]{Imp. (\%)} \\
			\midrule
			iCaRL \cite{Rebuffi_2017_CVPR} + FL & 82.0 & 80.0 & 67.0 & 62.0 & 61.3 & 60.3 & 57.0 & 54.3 & 53.0 & 51.7 & 50.3 & 50.0 & 48.7 & 48.0 & 46.7 & 45.0 & 45.0 & 44.0 & 43.3 & 42.7 & 54.6 & $\Uparrow$ 16.0 \\ 
			BiC \cite{wu2019large} + FL & 82.0 & 77.3 & 68.3 & 64.0 & 63.7 & 62.3 & 60.3 & 58.7 & 55.0 & 53.3 & 52.0 & 51.3 & 50.3 & 49.7 & 48.0 & 47.0 & 46.3 & 45.7 & 45.3 & 44.3 & 56.2 & $\Uparrow$ 14.4 \\ 
			PODNet \cite{10.1007/978-3-030-58565-5_6} + FL & 83.0 & 76.3 & 70.3 & 68.0 & 66.3 & 67.0 & 65.3 & 61.7 & 61.3 & 58.7 & 56.3 & 55.0 & 54.0 & 53.0 & 51.0 & 50.3 & 49.3 & 48.0 & 48.3 & 47.7 & 59.5 & $\Uparrow$ 11.1 \\ 
			DDE \cite{Hu_2021_CVPR} + iCaRL \cite{Rebuffi_2017_CVPR} + FL & 83.0 & 75.3 & 69.7 & 65.0 & 67.0 & 63.7 & 59.3 & 58.0 & 60.0 & 55.3 & 54.7 & 54.0 & 53.3 & 52.0 & 50.7 & 50.0 & 49.3 & 48.7 & 48.0 & 47.3 & 58.2 & $\Uparrow$ 12.4 \\
			GeoDL \cite{Christian2021MGeoCont} + iCaRL \cite{Rebuffi_2017_CVPR} + FL & 82.0& 78.3 & 71.3 & 67.7 & 68.0 & 65.3 & 64.3 & 60.0 & 58.7 &56.0 & 55.3 & 55.0 & 53.7 &53.0 & 51.7 & 50.7 & 50.0 & 49.0 & 49.3 & 48.0 & 59.4 & $\Uparrow$ 11.2 \\
			SS-IL \cite{Ahn_2021_ICCV} + FL & 83.0 & 73.3 & 63.7 & 61.3 & 60.3 & 59.3 & 57.3 & 56.0 & 54.7 & 53.3 &52.3 & 52.0 & 51.3 & 50.7 & 50.0 & 49.3 & 49.0 & 48.3 & 48.0 & 47.7 & 56.0 & $\Uparrow$ 14.6 \\		
			DyTox \cite{Douillard_2022_CVPR} + FL & 79.6& 78.3& 67.1&65.6& 68.5& 64.3& 63.7& 61.0& 58.8&59.0& 56.2& 58.5& 58.3& 58.2& 55.0& 51.8& 49.7 &48.7& 49.0& 52.7 & 60.2 & $\Uparrow$ 10.4 \\
			AFC \cite{Kang_2022_CVPR} + FL &75.6 & 69.6 & 57.1 &58.5& 45.5& 55.4& 51.4& 50.4& 45.2& 42.4&41.3& 35.6& 37.1& 37.8& 38.9& 35.2& 34.4& 34.5& 36.2& 33.8 & 45.8 & $\Uparrow$ 24.8 \\
			GLFC \cite{Dong_2022_CVPR} & 82.2& 82.5& 74.9& 75.2& 73.3& 71.5& 70.1& 67.7& 64.6& 65.9& 63.7& 64.2& 62.0& 61.0& 60.2& 58.9& 57.6& 59.3& 56.8& 56.8 & 66.4 & $\Uparrow$ 4.2 \\ 
			\midrule
			\midrule
			Ours-w/oCBL &81.6& 84.0& 77.9& 75.7& 75.2& 72.3& 71.3& 67.0& 64.6& 64.8& 63.2& 61.6& 58.7& 58.2& 55.6& 54.1& 52.3& 52.7& 50.1& 49.6 & 64.5 & $\Uparrow$ 6.1 \\
			Ours-w/oSDL &82.2& 80.2& 70.4& 67.0& 65.2& 61.4& 59.7& 57.8& 52.9& 56.0& 52.4& 51.5& 49.8& 49.7& 48.2& 46.9& 46.5& 44.3& 43.1& 41.9 & 56.4 & $\Uparrow$ 14.2 \\
			Ours-w/oPSR & 83.4& 85.4& 80.1& 76.8& 75.6& 73.9& 72.8& 69.4& 65.5& 65.3& 64.5& 62.9& 61.3& 60.9& 57.3& 56.9& 57.2& 58.0& 57.0& 56.8 & 67.1 & $\Uparrow$ 3.5 \\ 
			\textbf{Ours (LGA)} &\textcolor[rgb]{0.698,0.133,0.133}{\textbf{85.8}}& \textcolor[rgb]{0.698,0.133,0.133}{\textbf{85.9}}& \textcolor[rgb]{0.698,0.133,0.133}{\textbf{80.7}}& \textcolor[rgb]{0.698,0.133,0.133}{\textbf{78.9}}& \textcolor[rgb]{0.698,0.133,0.133}{\textbf{78.4}}& \textcolor[rgb]{0.698,0.133,0.133}{\textbf{74.6}}& \textcolor[rgb]{0.698,0.133,0.133}{\textbf{75.1}}& \textcolor[rgb]{0.698,0.133,0.133}{\textbf{71.3}}& \textcolor[rgb]{0.698,0.133,0.133}{\textbf{68.9}}& \textcolor[rgb]{0.698,0.133,0.133}{\textbf{69.2}}& \textcolor[rgb]{0.698,0.133,0.133}{\textbf{68.3}}& \textcolor[rgb]{0.698,0.133,0.133}{\textbf{67.7}}& \textcolor[rgb]{0.698,0.133,0.133}{\textbf{65.5}}& \textcolor[rgb]{0.698,0.133,0.133}{\textbf{65.6}}& \textcolor[rgb]{0.698,0.133,0.133}{\textbf{64.0}}& \textcolor[rgb]{0.698,0.133,0.133}{\textbf{63.0}}& \textcolor[rgb]{0.698,0.133,0.133}{\textbf{63.1}}& \textcolor[rgb]{0.698,0.133,0.133}{\textbf{63.7}}& \textcolor[rgb]{0.698,0.133,0.133}{\textbf{61.6}}& \textcolor[rgb]{0.698,0.133,0.133}{\textbf{60.5}} & \textcolor[rgb]{0.698,0.133,0.133}{\textbf{70.6}} & --- \\ 
			\bottomrule 
	\end{tabular}   } 
	\label{tab: exp_cifar100_task20}
\end{table*}

\begin{table*}[t]
	\centering
	\setlength{\tabcolsep}{0.9mm}
	\caption{Comparison experiments in terms of accuracy on MiniImageNet dataset \cite{NIPS2016_90e13578} when setting the number of consecutive learning tasks as $T=20$. }
	\scalebox{0.905}{
		\begin{tabular}{c|cccccccccccccccccccc|>{\columncolor{lightgray}}c|>{\columncolor{lightgray}}l}
			\toprule
			Methods & 5 & 10 & 15 & 20 & 25 & 30 & 35 & 40 & 45 & 50 & 55 & 60 & 65 & 70 & 75 & 80 & 85 & 90 & 95 & 100 & Avg. (\%) & \makecell[c]{Imp. (\%)} \\
			\midrule
			iCaRL \cite{Rebuffi_2017_CVPR} + FL & 83.0& 66.0& 61.3 & 56.0 & 56.3 & 53.0 & 49.7 & 47.0 & 46.3& 46.0 & 44.0 & 42.3 & 40.0 & 39.7 & 37.3 & 36.0 & 34.7 & 34.3 & 33.0 & 32.0 & 46.9 & $\Uparrow$ 16.0 \\ 
			BiC \cite{wu2019large} + FL & 82.3 & 64.7 & 59.0 & 58.3 & 57.0 & 54.7 & 52.3 & 50.3 & 49.0 & 47.7 & 46.7 & 44.0 & 42.7 & 41.3 & 40.3 & 38.0 & 37.0 & 36.3 & 34.7 & 33.0 & 48.5 & $\Uparrow$ 14.4 \\ 
			PODNet \cite{10.1007/978-3-030-58565-5_6} + FL & 81.7 & 63.3 & 60.3 & 59.3 & 58.3 & 56.3 & 55.0 & 53.3 & 51.7 & 50.0 & 49.3 & 48.0 & 47.0 & 45.3 & 44.7 & 43.7 & 42.0 & 39.7 & 38.7 & 37.0 & 51.2 & $\Uparrow$ 11.7 \\ 
			DDE \cite{Hu_2021_CVPR} + iCaRL \cite{Rebuffi_2017_CVPR} + FL & 80.0 & 60.7 & 58.7 & 56.3 & 57.0 & 55.3 & 53.0 & 51.7 & 50.3 & 49.3 & 48.7 & 48.3 & 47.7 & 46.7 & 45.7 & 44.3 & 42.3 & 40.0 & 38.3 & 37.3 & 50.6 & $\Uparrow$ 12.3 \\
			GeoDL \cite{Christian2021MGeoCont} + iCaRL \cite{Rebuffi_2017_CVPR} + FL & 82.3 & 66.3 & 62.7 & 61.0 & 60.3 & 58.0 & 56.3 & 55.3 & 53.0 & 51.3 & 50.0 & 48.7 & 48.0 & 46.3 & 45.0 & 44.0 & 41.7 & 40.0 & 38.0 & 36.7 & 52.2 & $\Uparrow$ 10.7 \\
			SS-IL \cite{Ahn_2021_ICCV} + FL & 80.0 & 65.3 & 61.7 & 57.3 & 56.3 & 54.0 &51.3 & 50.0 & 49.3 & 48.3 & 47.0 & 45.0 & 44.3 & 43.0 & 41.3 & 40.7 & 39.3 & 38.7 & 37.0 & 36.0 & 49.3 & $\Uparrow$ 13.6 \\		
			DyTox \cite{Douillard_2022_CVPR} + FL & 71.6 &52.7& 61.6& 53.2& 56.8& 48.9& 45.7& 49.4& 39.1& 44.1& 37.7 & 35.2 & 33.6 & 31.5 & 28.6 & 27.3 & 27.1 & 26.5 & 25.8 & 24.9 & 41.1 & $\Uparrow$ 21.8 \\
			AFC \cite{Kang_2022_CVPR} + FL & 72.4& 53.0& 51.8& 38.1& 41.4 & 39.6& 41.2& 37.2& 33.7& 32.4& 29.7& 33.5& 29.6& 30.2& 25.1& 25.1& 26.1& 24.6& 24.0& 23.5 & 35.6 & $\Uparrow$ 27.3 \\
			GLFC \cite{Dong_2022_CVPR} & \textcolor[rgb]{0.698,0.133,0.133}{\textbf{84.0}} & 71.7 & 70.0 &69.3 &67.3 & 66.3 & 61.0 & 60.7 & 59.3 & 58.7 & 55.3 & 53.0 & 52.0 & 50.3 &49.7 & 47.3 & 46.0 & 42.7 & 40.3 & 39.0 & 57.2 & $\Uparrow$ 5.7 \\ 
			\midrule
			\midrule
			Ours-w/oCBL & 80.6& 78.5& 72.3& 67.3& 63.4& 63.5& 62.1& 64.2& 61.0& 59.0& 57.6& 58.4& 56.9& 53.8& 53.5& 51.5& 51.7& 50.6& 49.5& 45.6 & 60.1 & $\Uparrow$ 2.8 \\
			Ours-w/oSDL &72.2& 64.5& 70.1& 69.7& 65.7& 52.0& 53.8& 52.8& 50.9& 51.0& 45.9& 45.6& 42.5& 41.5& 40.8& 39.0& 38.9& 36.2& 36.1& 35.7 & 50.2 & $\Uparrow$ 12.7 \\
			Ours-w/oPSR &79.0& 78.3& 74.9& 71.0 & 67.4 & 65.8 & 65.1 & 64.7 &62.1 & 60.9 & 58.2 & 57.4 & 55.6 & 54.2 & 53.8 & 53.1 & 52.3 & 51.8 & 50.4 & 46.3 & 61.1 & $\Uparrow$ 1.8 \\
			
			\textbf{Ours (LGA)} &78.8 & \textcolor[rgb]{0.698,0.133,0.133}{\textbf{79.4}}& \textcolor[rgb]{0.698,0.133,0.133}{\textbf{76.8}} & \textcolor[rgb]{0.698,0.133,0.133}{\textbf{73.5}} & \textcolor[rgb]{0.698,0.133,0.133}{\textbf{69.8}} & \textcolor[rgb]{0.698,0.133,0.133}{\textbf{68.5}} & \textcolor[rgb]{0.698,0.133,0.133}{\textbf{67.3}} & \textcolor[rgb]{0.698,0.133,0.133}{\textbf{66.1}} & \textcolor[rgb]{0.698,0.133,0.133}{\textbf{63.8}} & \textcolor[rgb]{0.698,0.133,0.133}{\textbf{62.1}} & \textcolor[rgb]{0.698,0.133,0.133}{\textbf{60.6}} & \textcolor[rgb]{0.698,0.133,0.133}{\textbf{59.8}} & \textcolor[rgb]{0.698,0.133,0.133}{\textbf{57.2}} & \textcolor[rgb]{0.698,0.133,0.133}{\textbf{56.8}} & \textcolor[rgb]{0.698,0.133,0.133}{\textbf{55.1}} & \textcolor[rgb]{0.698,0.133,0.133}{\textbf{54.7}} & \textcolor[rgb]{0.698,0.133,0.133}{\textbf{54.1}} & \textcolor[rgb]{0.698,0.133,0.133}{\textbf{53.2}} & \textcolor[rgb]{0.698,0.133,0.133}{\textbf{51.6}} & \textcolor[rgb]{0.698,0.133,0.133}{\textbf{48.2}} & \textcolor[rgb]{0.698,0.133,0.133}{\textbf{62.9}} & --- \\ 
			\bottomrule 
	\end{tabular}   } 
	\label{tab: exp_imagenetsubset_task20}
\end{table*}

\begin{table*}[t]
	\centering
	\setlength{\tabcolsep}{0.9mm}
	\caption{Comparison experiments in terms of accuracy on TinyImageNet dataset \cite{Tiny_imagenet} when setting the number of consecutive learning tasks as $T=20$. }
	\scalebox{0.905}{
		\begin{tabular}{c|cccccccccccccccccccc|>{\columncolor{lightgray}}c|>{\columncolor{lightgray}}l}
			\toprule
			Methods & 10 & 20 & 30 & 40 & 50 & 60 & 70 & 80 & 90 & 100 & 110 & 120 & 130 & 140 & 150 & 160 & 170 & 180 & 190 & 200 & Avg. (\%) & \makecell[c]{Imp. (\%)} \\
			\midrule
			iCaRL \cite{Rebuffi_2017_CVPR} + FL & 67.0 & 59.3 & 54.0 & 48.3 & 46.7 & 44.7 & 43.3 & 39.0 & 37.3 & 33.0 & 32.0 & 30.3 & 28.0 & 27.0 & 26.3 & 25.3 & 24.7 & 24.0 & 22.7 & 22.0 & 36.7 & $\Uparrow$ 12.4 \\ 
			BiC \cite{wu2019large} + FL & 67.3 & 59.7 & 54.7 & 50.0 & 48.3 & 45.3 & 43.0 & 40.7 & 38.0 & 33.7 & 32.7 & 32.3 & 30.3 & 29.0 & 27.7 & 27.3 & 26.0 & 25.7 & 24.3 & 23.3 & 38.0 & $\Uparrow$ 11.1 \\ 
			PODNet \cite{10.1007/978-3-030-58565-5_6} + FL & 69.0 & 59.3 & 55.0 & 51.7 & 50.0 & 46.7 & 43.7 & 41.0 & 39.3 & 38.0 & 37.0 & 35.7 & 34.7 & 34.0 & 33.0 & 32.3 & 31.0 & 30.0 & 29.3 & 28.0 & 40.9 & $\Uparrow$ 8.2 \\ 
			DDE \cite{Hu_2021_CVPR} + iCaRL \cite{Rebuffi_2017_CVPR} + FL & 70.0 & 59.3 & 53.3& 51.0 & 48.3 & 45.7 & 42.3 & 40.0 & 38.0 & 36.3 & 35.0 & 33.7 & 32.0 & 31.0 & 30.3 & 30.0 & 28.7 & 28.3 & 27.3 & 26.0 & 39.3 & $\Uparrow$ 9.8 \\
			GeoDL \cite{Christian2021MGeoCont} + iCaRL \cite{Rebuffi_2017_CVPR} + FL & 66.3 & 56.7 & 51.0 & 49.7 & 44.7 & 42.3 & 41.0 & 39.0 & 37.3 & 35.0 & 33.7 & 32.0 & 31.0 & 30.3 & 28.7 & 28.0 & 27.3 & 26.3 & 25.0 & 24.7 & 37.5 & $\Uparrow$ 11.6 \\
			SS-IL \cite{Ahn_2021_ICCV} + FL & 66.7 & 54.0 & 47.7 & 45.3 & 42.3 & 42.0 & 40.7 & 38.0 & 36.0 & 34.3 & 33.0 & 31.0 & 29.3 & 28.3 & 27.7 & 27.0 & 26.3 & 26.0 & 25.0 & 24.3 & 36.2 & $\Uparrow$ 12.9 \\		
			DyTox \cite{Douillard_2022_CVPR} + FL & \textcolor[rgb]{0.698,0.133,0.133}{\textbf{77.6}}& \textcolor[rgb]{0.698,0.133,0.133}{\textbf{70.2}}& 63.4& 56.6& 52.0& 44.6& 51.6& 39.6& 41.5& 39.0& 37.8& 31.2& 34.2& 30.6 & 29.8 & 29.2 & 28.3 & 27.5 & 26.8 & 15.3 & 41.3 & $\Uparrow$ 7.8 \\
			AFC \cite{Kang_2022_CVPR} + FL & 74.0& 62.9& 57.6& 54.2& 45.1& 44.4& 40.7& 36.9& 33.0& 33.6& 30.8& 28.9& 27.1& 22.8& 24.5& 23.6& 22.1& 20.7& 18.4& 18.1 & 36.0 & $\Uparrow$ 13.1 \\
			GLFC \cite{Dong_2022_CVPR} & 68.7 & 63.3 & 61.7 & 57.3 & 56.0 & 53.0 & 50.3 & 47.7 & 46.3 & 45.0 & 42.7 & 41.0 & 40.0 & 39.3 & 38.0 & 36.7 & 35.3 & 34.0 & 33.0 & 31.7 & 46.1 & $\Uparrow$ 3.0 \\
			\midrule
			\midrule
			Ours-w/oCBL & 73.6& 67.3& 62.4& 57.4& 51.1& 49.6& 47.0& 43.6& 39.4& 37.3& 34.6& 30.9& 28.7& 17.2 & 18.5 & 15.4 & 14.7 & 13.5 & 12.4 & 11.7 & 36.3 & $\Uparrow$ 12.8 \\
			Ours-w/oSDL & 73.8& 47.0& 53.1& 51.1& 47.2& 46.4& 44.2& 42.1& 38.6& 38.0& 36.1& 35.0& 33.1& 30.2& 30.1& 28.1& 27.5& 26.0& 24.1& 23.4 & 38.8 & $\Uparrow$ 10.3 \\
			Ours-w/oPSR & 72.4& 68.0& 59.5& 60.3& 51.0& 50.4 & 48.6 & 45.7 & 42.1 & 39.7 & 36.3 & 35.9 & 33.5 & 32.7 & 33.8 & 31.9 & 30.5 & 28.8 & 28.1 & 27.4 & 42.8 & $\Uparrow$ 6.3 \\
			\textbf{Ours (LGA)} &74.0 & 67.6& \textcolor[rgb]{0.698,0.133,0.133}{\textbf{64.9}}& \textcolor[rgb]{0.698,0.133,0.133}{\textbf{61.0}} & \textcolor[rgb]{0.698,0.133,0.133}{\textbf{58.9}} & \textcolor[rgb]{0.698,0.133,0.133}{\textbf{55.7}} & \textcolor[rgb]{0.698,0.133,0.133}{\textbf{53.6}} & \textcolor[rgb]{0.698,0.133,0.133}{\textbf{51.3}} & \textcolor[rgb]{0.698,0.133,0.133}{\textbf{50.1}} & \textcolor[rgb]{0.698,0.133,0.133}{\textbf{48.8}} & \textcolor[rgb]{0.698,0.133,0.133}{\textbf{45.2}} & \textcolor[rgb]{0.698,0.133,0.133}{\textbf{43.7}} & \textcolor[rgb]{0.698,0.133,0.133}{\textbf{42.8}} & \textcolor[rgb]{0.698,0.133,0.133}{\textbf{41.2}} & \textcolor[rgb]{0.698,0.133,0.133}{\textbf{40.5}} & \textcolor[rgb]{0.698,0.133,0.133}{\textbf{38.9}} & \textcolor[rgb]{0.698,0.133,0.133}{\textbf{37.4}} & \textcolor[rgb]{0.698,0.133,0.133}{\textbf{36.6}} & \textcolor[rgb]{0.698,0.133,0.133}{\textbf{35.1}} & \textcolor[rgb]{0.698,0.133,0.133}{\textbf{33.8}} & \textcolor[rgb]{0.698,0.133,0.133}{\textbf{49.1}} & --- \\ 
			\bottomrule
	\end{tabular}	} 	
	\label{tab: exp_tinyimageNet_task20}
\end{table*}

\subsection{Implementation Details}
In the FCIL settings, as introduced in \cite{Dong_2022_CVPR}, we follow the identical protocols proposed in \cite{Rebuffi_2017_CVPR, wu2019large} to set a series of consecutive learning tasks for fair comparison experiments with some representative class-incremental learning (CIL) models \cite{Rebuffi_2017_CVPR, wu2019large, 10.1007/978-3-030-58565-5_6, Hu_2021_CVPR, Christian2021MGeoCont, Ahn_2021_ICCV, Douillard_2022_CVPR, Kang_2022_CVPR}. Specifically, we employ the same class order used in iCaRL \cite{Rebuffi_2017_CVPR}, set $T=\{5, 10, 20\}$ to perform different incremental tasks on benchmark datasets, and utilize the same classification backbone ResNet-18 \cite{7780459} as local model. Following baseline iCaRL \cite{Rebuffi_2017_CVPR}, we use the same strategies (\emph{i.e.}, random horizontal flip, color jitter, normalization and random cropping) for data augmentation.  Besides, the mixup technology proposed in DyTox \cite{Douillard_2022_CVPR} is not used in this paper. We set the exemplar memory $\mathcal{M}_l$ as $2,000$, and follow iCaRL \cite{Rebuffi_2017_CVPR} to update $\mathcal{M}_l$ as the incremental tasks. The buffer size of exemplar set $\mathcal{M}_l$ is fixed as new classes increase. For the $c$-th ($c=1, \cdots, C_l^o+C_l^t$) class, the $l$-th local client can store the top $\frac{|\mathcal{M}_l|}{C_l^o+C_l^t}$ exemplars whose features are closest to the average embedding of all samples belonging to the $c$-th class. Thus, we store $\frac{|\mathcal{M}_l|}{C_l^o+C_l^t}$ exemplars per new class and remove $\frac{|\mathcal{M}_l|}{C_l^o}-\frac{|\mathcal{M}_l|}{C_l^o+C_l^t}$ samples per old class in the $t$-th task.

All local models are optimized via the SGD optimizer with initial learning rate as 2.0. We consider a 4-layer LeNet \cite{726791} as the gradient encoding network $\Gamma$ in this paper. Moreover, local clients utilize a SGD optimizer to generate representative perturbed images, and proxy server employs a L-BFGS optimizer to reconstruct prototype images, where their learning rates are initialized as $1.0$. 
In the first learning task, the number of local clients is set as $30$. As the consecutive learning tasks arrive, we add $10$ new local clients for each new task. In the FCIL settings, $10$ local clients are chosen to perform $20$-epoch local training at each global round. Considering Non-IID class imbalance across different local clients, we randomly assign $60\%$ categories from the label space of current learning task to local clients for CIFAR-100 \cite{krizhevsky2009learning}, MiniImageNet \cite{NIPS2016_90e13578} and TinyImageNet \cite{Tiny_imagenet}, while we set $70\%$ for ImageNet-1000 \cite{5206848}. 
All experiment results shown in this paper are averaged over 3 random runs, where random seeds are set as $\{2021, 2022, 2023\}$.

\subsection{Comparison Experiments}
As shown in Tables~\ref{tab: exp_all_datasets_task5}$\sim$\ref{tab: exp_tinyimageNet_task20}, we introduce a large number of comparison experiments between our model and other competing methods on CIFAR-100 \cite{krizhevsky2009learning}, MiniImageNet \cite{NIPS2016_90e13578}, TinyImageNet \cite{Tiny_imagenet} and ImageNet-1000 \cite{5206848} datasets when setting the number of consecutive learning tasks as $T=\{5, 10, 20\}$. Avg. (\%) is the averaged performance of different tasks, and Imp. (\%) denotes the performance improvement of our model over other comparison methods. We have the following conclusions from the presented results in Tables~\ref{tab: exp_all_datasets_task5}$\sim$\ref{tab: exp_tinyimageNet_task20}: 1) The proposed LGA model has significant performance improvements ($2.9\%\sim10.5\%$ in terms of averaged accuracy) over our conference version (\emph{i.e.}, GLFC \cite{Dong_2022_CVPR}), since the proposed category-balanced gradient-adaptive compensation loss $\mathcal{L}_{\mr{CB}}$ and category gradient-induced semantic distillation loss $\mathcal{L}_{\mr{SD}}$ can effectively perform local anti-forgetting on old classes. 2) When compared with other baseline class-incremental learning (CIL) methods \cite{Rebuffi_2017_CVPR, wu2019large, 10.1007/978-3-030-58565-5_6, Hu_2021_CVPR, Christian2021MGeoCont, Ahn_2021_ICCV, Douillard_2022_CVPR, Kang_2022_CVPR}, our model outperforms them about $4.8\%\sim27.3\%$ averaged accuracy. It validates the effectiveness of the improved proxy server to perform global anti-forgetting on old classes via selecting the best old model from a global perspective. 3) When performing on different kinds of consecutive learning tasks (\emph{i.e.}, $T=\{5, 10, 20\}$), our LGA model has achieved state-of-the-art performance against other comparison methods, which illustrates the robustness and effectiveness of our LGA model to tackle local and global forgetting under different experimental settings in the FCIL.

\begin{figure*}[t]
	\centering
	\includegraphics[width=520pt, height=250pt]
	{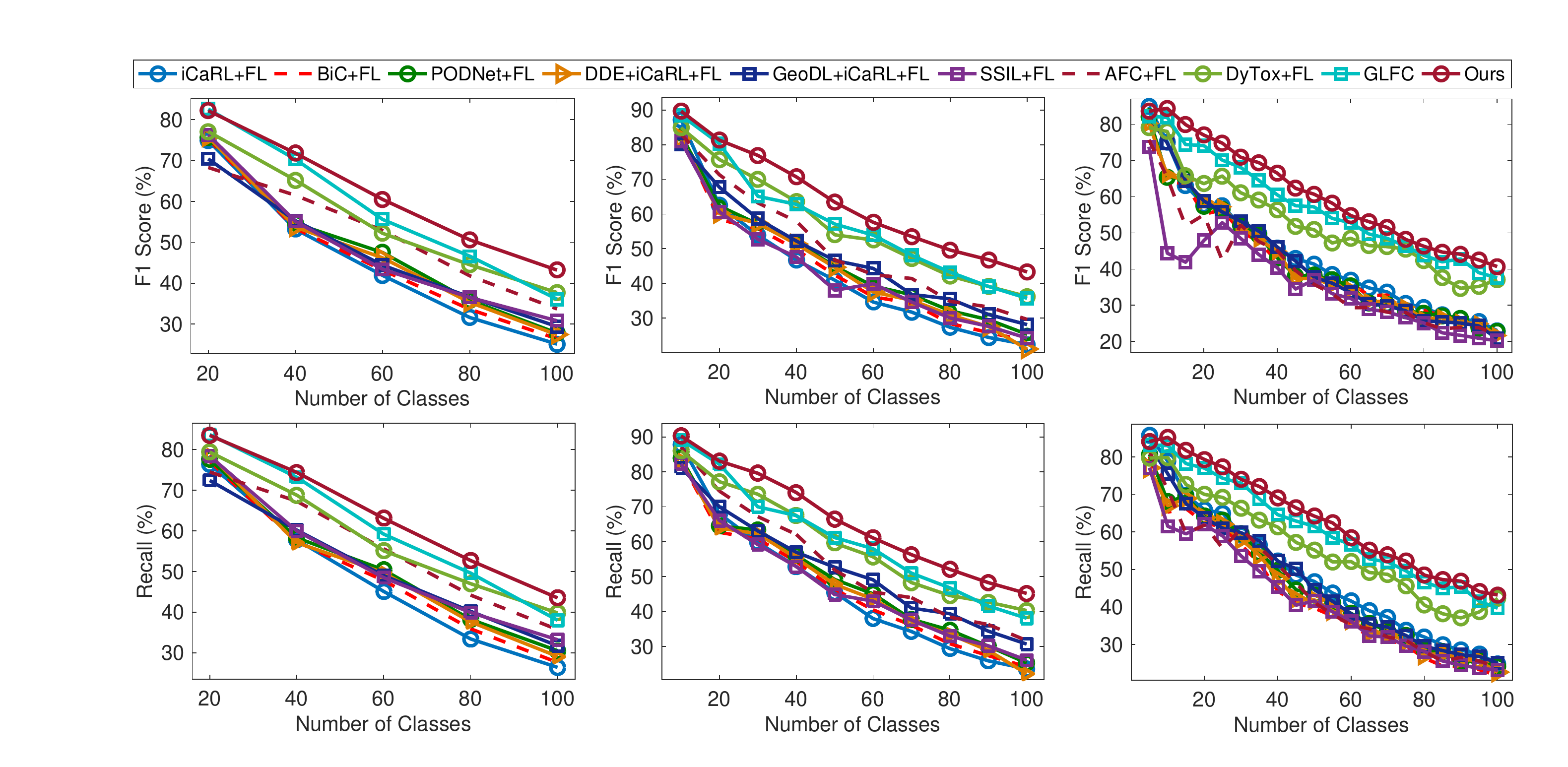}
	\vspace{-20pt}
	\caption{Investigation of anti-forgetting in terms of F1 score and recall on CIFAR-100 \cite{krizhevsky2009learning} when $T=5$ (left), $T=10$ (middle) and $T=20$ (right). }
	\label{fig: incremental_tasks_CIFAR100_F1}
\end{figure*}
\begin{figure*}[t]
	\centering
	\includegraphics[width=520pt, height=250pt]
	{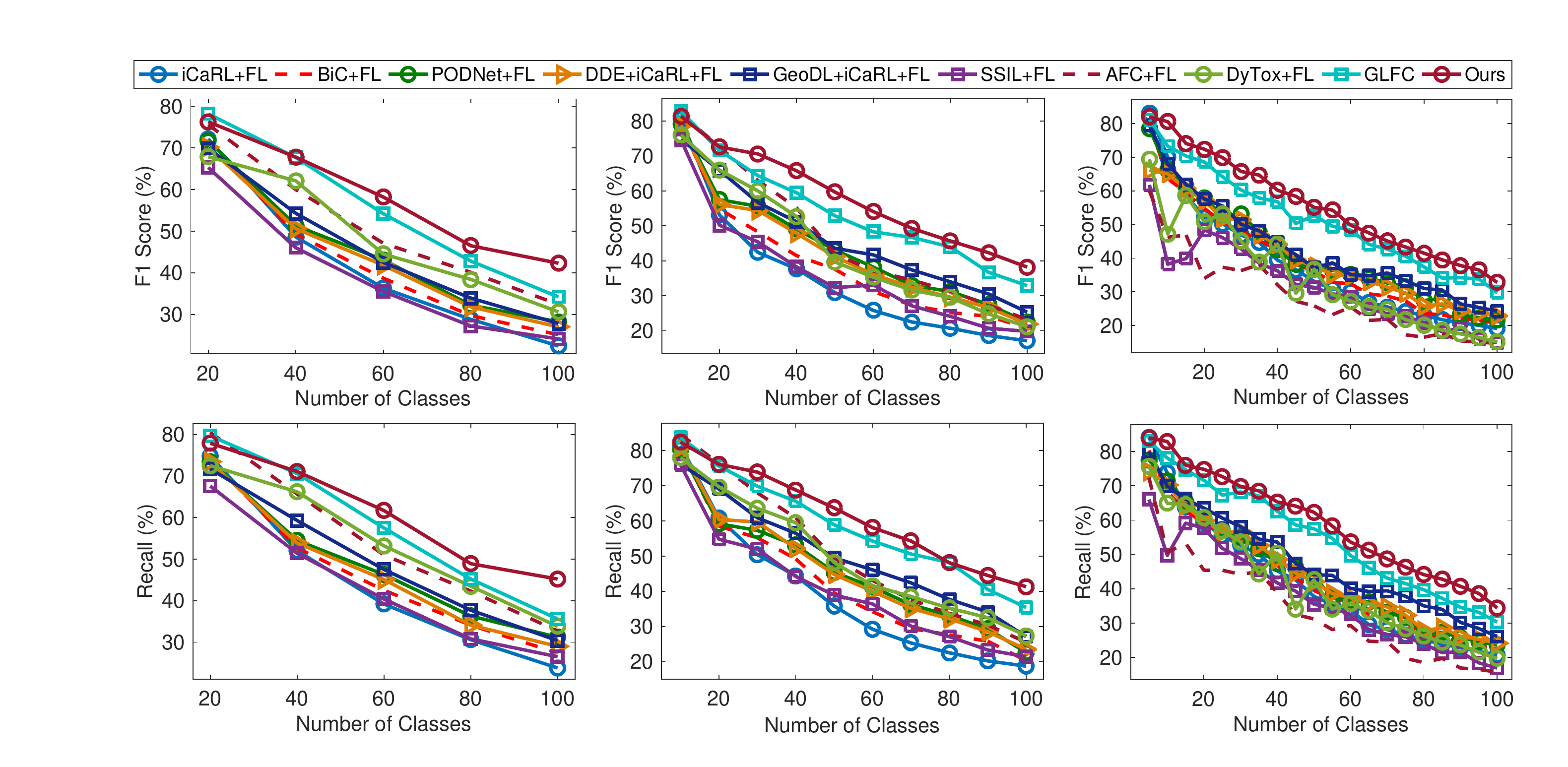}
	\vspace{-20pt}
	\caption{Investigation of anti-forgetting in terms of F1 score and recall on MiniImageNet \cite{NIPS2016_90e13578} when $T=5$ (left), $T=10$ (middle) and $T=20$ (right). }
	\label{fig: incremental_tasks_ImageNetSub_F1}
\end{figure*}

\subsection{Ablation Studies}
This subsection shows ablation experiments to illustrate the effectiveness of each proposed module in our LGA model under different settings (\emph{i.e.}, $T=\{5, 10, 20\}$), as shown in Tables~\ref{tab: exp_all_datasets_task5}$\sim$\ref{tab: exp_tinyimageNet_task20}. Ours-w/oCBL, Ours-w/oSDL and Ours-w/oPSR represent the performance of training proposed LGA model without utilizing the category-balanced gradient-adaptive compensation loss $\mathcal{L}_{\mr{CB}}$ (CBL), category gradient-induced semantic distillation loss $\mathcal{L}_{\mr{SD}}$ (SDL) and proxy server $\mathcal{S}_p$ (PSR). Ours-w/oCBL and Ours-w/oSDL use cross-entropy loss $\mathcal{L}_{\mr{CE}}$ and distillation loss proposed in iCaRL \cite{Rebuffi_2017_CVPR} for substitution. Ours-w/oPSR cannot use proxy server to choose the best old global model for global anti-forgetting, and it randomly selects an old global model for semantic distillation in ablation studies.

\begin{table*}[t]
	\centering
	\setlength{\tabcolsep}{0.72mm}
	\caption{Investigation of heterogeneous class distribution ($\{40\%, 50\%, 60\%\}$) in terms of accuracy and F1 score on CIFAR-100 \cite{krizhevsky2009learning} when setting $T=10$. }
	\scalebox{0.9}{
		\begin{tabular}{c|c|cccccccccc|>{\columncolor{lightgray}}c|>{\columncolor{lightgray}}c|cccccccccc|>{\columncolor{lightgray}}c>{\columncolor{lightgray}}c}
			\toprule
			\multirow{2}{*}{} & \multirow{2}{*}{Methods} & \multicolumn{12}{c|}{Accuracy} & \multicolumn{12}{c}{F1 Score} \\
			& & 10 & 20 & 30 & 40 & 50 & 60 & 70 & 80 & 90 & 100 & Avg. (\%) & Imp. (\%) & 10 & 20 & 30 & 40 & 50 & 60 & 70 & 80 & 90 & 100 & Avg. (\%) & Imp. (\%) \\
			\midrule
			\multicolumn{1}{c|}{\multirow{2}{*}{40\%}} & GLFC \cite{Dong_2022_CVPR} & 74.9& 64.9& \textcolor[rgb]{0.698,0.133,0.133}{\textbf{66.0}}& \textcolor[rgb]{0.698,0.133,0.133}{\textbf{64.8}}& 58.2& 58.7& 58.1& 50.1& 49.0& 49.2 & 59.4 & $\Uparrow$ 3.6 & 72.2& 59.4& \textcolor[rgb]{0.698,0.133,0.133}{\textbf{60.6}}& \textcolor[rgb]{0.698,0.133,0.133}{\textbf{58.6}}& 48.4& 47.8& 44.8& 36.0& 32.8& 32.2 & 49.3 & $\Uparrow$ 3.6 \\ 
			\multicolumn{1}{c|}{} & \textbf{Ours (LGA)} & \textcolor[rgb]{0.698,0.133,0.133}{\textbf{77.4}}& \textcolor[rgb]{0.698,0.133,0.133}{\textbf{70.1}}& 65.5& 63.8 & \textcolor[rgb]{0.698,0.133,0.133}{\textbf{61.4}}& \textcolor[rgb]{0.698,0.133,0.133}{\textbf{61.3}}& \textcolor[rgb]{0.698,0.133,0.133}{\textbf{61.8}}& \textcolor[rgb]{0.698,0.133,0.133}{\textbf{57.1}}& \textcolor[rgb]{0.698,0.133,0.133}{\textbf{56.8}}& \textcolor[rgb]{0.698,0.133,0.133}{\textbf{55.1}} & \textcolor[rgb]{0.698,0.133,0.133}{\textbf{63.0}} & --- & \textcolor[rgb]{0.698,0.133,0.133}{\textbf{75.1}}& \textcolor[rgb]{0.698,0.133,0.133}{\textbf{65.6}}& 59.1&57.1& \textcolor[rgb]{0.698,0.133,0.133}{\textbf{52.2}}& \textcolor[rgb]{0.698,0.133,0.133}{\textbf{51.1}}& \textcolor[rgb]{0.698,0.133,0.133}{\textbf{49.1}}& \textcolor[rgb]{0.698,0.133,0.133}{\textbf{42.2}}& \textcolor[rgb]{0.698,0.133,0.133}{\textbf{40.1}}& \textcolor[rgb]{0.698,0.133,0.133}{\textbf{37.0}} & \textcolor[rgb]{0.698,0.133,0.133}{\textbf{52.9}} & --- \\
			\midrule
			
			\multicolumn{1}{c|}{\multirow{2}{*}{50\%}} & GLFC \cite{Dong_2022_CVPR} & 86.2& 64.0& 71.3& 65.7& 65.3& 61.5& 60.7 & 57.8 & 58.9& 50.3 & 64.2 & $\Uparrow$ 6.0 & 85.9& 58.9 &65.9& 61.0& 56.6& 50.3& 47.6& 41.2& 42.0& 32.5 & 54.2 & $\Uparrow$ 6.4 \\ 
			\multicolumn{1}{c|}{} & \textbf{Ours (LGA)} & \textcolor[rgb]{0.698,0.133,0.133}{\textbf{86.3}}& \textcolor[rgb]{0.698,0.133,0.133}{\textbf{78.5}}& \textcolor[rgb]{0.698,0.133,0.133}{\textbf{74.7}}& \textcolor[rgb]{0.698,0.133,0.133}{\textbf{72.5}}& \textcolor[rgb]{0.698,0.133,0.133}{\textbf{70.7}}& \textcolor[rgb]{0.698,0.133,0.133}{\textbf{67.3}}& \textcolor[rgb]{0.698,0.133,0.133}{\textbf{65.7}}& \textcolor[rgb]{0.698,0.133,0.133}{\textbf{63.7}}& \textcolor[rgb]{0.698,0.133,0.133}{\textbf{62.9}}& \textcolor[rgb]{0.698,0.133,0.133}{\textbf{60.1}} & \textcolor[rgb]{0.698,0.133,0.133}{\textbf{70.2}} & ---  & \textcolor[rgb]{0.698,0.133,0.133}{\textbf{86.2}}& \textcolor[rgb]{0.698,0.133,0.133}{\textbf{76.8}}& \textcolor[rgb]{0.698,0.133,0.133}{\textbf{69.7}}& \textcolor[rgb]{0.698,0.133,0.133}{\textbf{66.5}}& \textcolor[rgb]{0.698,0.133,0.133}{\textbf{62.9}}& \textcolor[rgb]{0.698,0.133,0.133}{\textbf{56.5}}& \textcolor[rgb]{0.698,0.133,0.133}{\textbf{52.5}}& \textcolor[rgb]{0.698,0.133,0.133}{\textbf{48.4}}& \textcolor[rgb]{0.698,0.133,0.133}{\textbf{45.7}}& \textcolor[rgb]{0.698,0.133,0.133}{\textbf{41.1}} & \textcolor[rgb]{0.698,0.133,0.133}{\textbf{60.6}} & --- \\
			\midrule
			
			\multicolumn{1}{c|}{\multirow{2}{*}{60\%}} & GLFC \cite{Dong_2022_CVPR} & \textcolor[rgb]{0.698,0.133,0.133}{\textbf{90.0}} & 82.3 & 77.0 & 72.3 & 65.0 & 66.3 & 59.7 & 56.3 & 50.3 & 50.0 & 66.9 & $\Uparrow$ 6.6 & 88.5& 80.3& 65.1& 62.8& 57.1& 53.9& 48.1& 43.2& 39.1& 35.7 & 57.4 & $\Uparrow$ 6.5 \\ 
			\multicolumn{1}{c|}{} & \textbf{Ours (LGA)} & 89.6& \textcolor[rgb]{0.698,0.133,0.133}{\textbf{83.2}}& \textcolor[rgb]{0.698,0.133,0.133}{\textbf{79.3}}& \textcolor[rgb]{0.698,0.133,0.133}{\textbf{76.1}}& \textcolor[rgb]{0.698,0.133,0.133}{\textbf{72.9}}& \textcolor[rgb]{0.698,0.133,0.133}{\textbf{71.7}}& \textcolor[rgb]{0.698,0.133,0.133}{\textbf{68.4}}& 
			\textcolor[rgb]{0.698,0.133,0.133}{\textbf{65.7}}& \textcolor[rgb]{0.698,0.133,0.133}{\textbf{64.7}}& \textcolor[rgb]{0.698,0.133,0.133}{\textbf{62.9}} & \textcolor[rgb]{0.698,0.133,0.133}{\textbf{73.5}} & --- &\textcolor[rgb]{0.698,0.133,0.133}{\textbf{89.2}}& \textcolor[rgb]{0.698,0.133,0.133}{\textbf{81.7}}& \textcolor[rgb]{0.698,0.133,0.133}{\textbf{76.1}}& \textcolor[rgb]{0.698,0.133,0.133}{\textbf{70.9}}& \textcolor[rgb]{0.698,0.133,0.133}{\textbf{65.0}}& \textcolor[rgb]{0.698,0.133,0.133}{\textbf{61.0}}& \textcolor[rgb]{0.698,0.133,0.133}{\textbf{55.5}}& \textcolor[rgb]{0.698,0.133,0.133}{\textbf{49.9}}& \textcolor[rgb]{0.698,0.133,0.133}{\textbf{46.4}}& \textcolor[rgb]{0.698,0.133,0.133}{\textbf{42.9}}& \textcolor[rgb]{0.698,0.133,0.133}{\textbf{63.9}} & --- \\ 
			
			\bottomrule
			
		\end{tabular}
	} 	
	\label{tab: heterogeneous_cifar100_tasks10}
\end{table*}

\begin{figure}[t]
	\centering
	\includegraphics[width=0.485\textwidth]
	{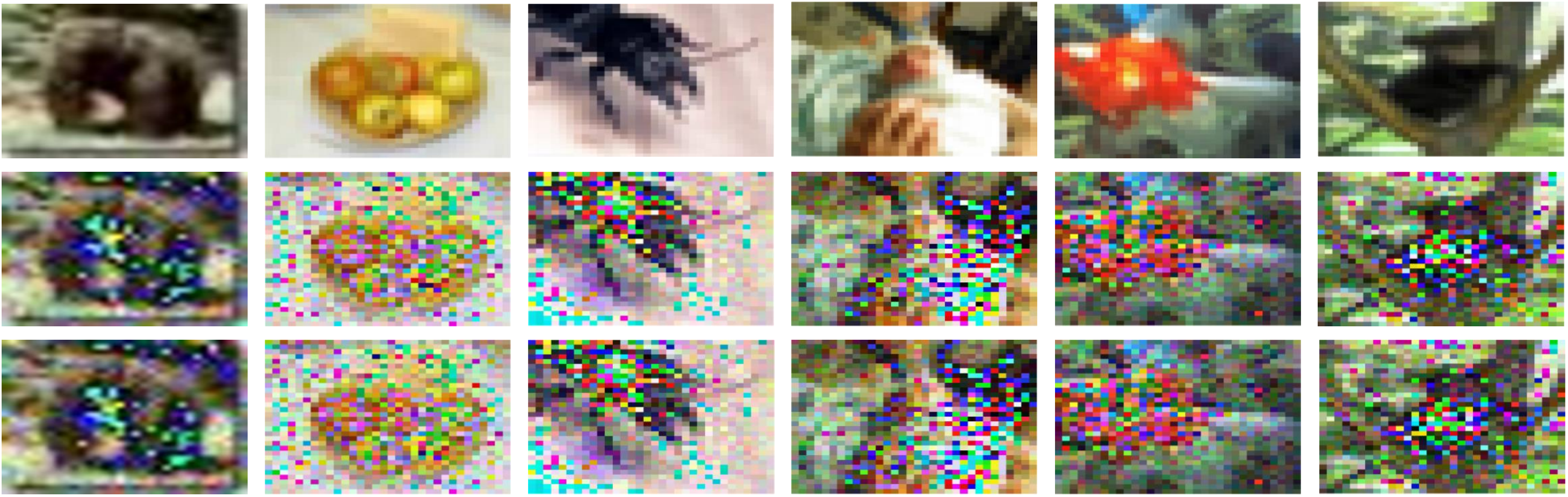}
	\vspace{-20pt}
	\caption{Investigation of privacy preservation on CIFAR-100 dataset \cite{krizhevsky2009learning}, where the top, middle and bottom rows denote the original prototype images, perturbed prototype images from local clients, and reconstructed prototype images of proxy server.  }
	\label{fig: prototype_samples}
\end{figure}

\begin{table}[t]
\centering
\setlength{\tabcolsep}{1.5mm}
\caption{Analysis about communication cost of proxy server. }
\scalebox{0.955}{
    \begin{tabular}{c|c|c}
        \toprule
        Methods & Communication Cost & Ratio \\
        \midrule
        Traditional FL \cite{DBLP:journals/corr/McMahanMRA16} & 4536.8MB & -- \\ 
    Proxy Server (\textbf{Ours}) & 16.5MB & 0.364\% \\ 
        \bottomrule
\end{tabular}	} 	
\label{tab: communication_cost}
\end{table}

Ours-w/oCBL decreases 2.8\%$\sim$12.8\% averaged accuracy compared with Ours. It validates the effectiveness of our model to balance heterogeneous forgetting speeds of hard-to-forget and easy-to-forget old categories via category-balanced gradient propagation. Besides, Ours significantly outperforms Ours-w/oSDL by a large margin about 7.3\%$\sim$18.3\% averaged accuracy. It demonstrates that $\mathcal{L}_{\mr{SD}}$ can effectively distill underlying inter-class relations within each task from the best old model to tackle local forgetting on old classes. 
From the ablation results in Tables~\ref{tab: exp_all_datasets_task5}$\sim$\ref{tab: exp_tinyimageNet_task20}, we can conclude that the SDL module has more performance improvement than CBL module under most FCIL settings. Such large improvement is caused by the distillation of semantic relations between old and new categories explored via the SDL module. Undoubtedly, compared with the CBL module, the inter-class semantic relations play a more important role in tackling local catastrophic forgetting on hard-to-forget and easy-to-forget old classes. Distilling inter-class relations between old and new models is more difficult under the real-world FCIL settings, and is more essential to balance heterogeneous forgetting speeds of hard-to-forget and easy-to-forget old categories, thus improving performance better than the CBL module. 
Moreover, the performance of Ours-w/oPSR decreases 0.9\%$\sim$8.8\% averaged accuracy. It validates the importance to address global catastrophic forgetting on old categories via the proposed proxy server.

\subsection{Analysis of Anti-Forgetting on Old Categories}
This subsection investigates the superiority of our LGA model to perform anti-forgetting on old categories when performing different kinds of incremental tasks ($T=\{5,10,20\}$) on CIFAR-100 \cite{krizhevsky2009learning} and MiniImageNet \cite{NIPS2016_90e13578} datasets. As depicted in Figs.~\ref{fig: incremental_tasks_CIFAR100_F1} and \ref{fig: incremental_tasks_ImageNetSub_F1}, we utilize F1 score and recall metrics to evaluate the anti-forgetting performance on old categories by predicting all classes instead of only the recently-seen classes. From the depicted curves in Figs.~\ref{fig: incremental_tasks_CIFAR100_F1} and \ref{fig: incremental_tasks_ImageNetSub_F1}, we observe that our LGA model performs better than the conference version (\emph{i.e.}, GLFC \cite{Dong_2022_CVPR}) to achieve anti-forgetting on old categories across different learning tasks. When compared with other class-incremental learning (CIL) methods \cite{Rebuffi_2017_CVPR, wu2019large, 10.1007/978-3-030-58565-5_6, Douillard_2022_CVPR, Kang_2022_CVPR}, our model significantly outperforms them in terms of F1 score and recall via designing proxy server to choose the best old model for local training. It illustrates the effectiveness of our proposed LGA model to accurately identify both old and new categories instead of only the recently-seen novel classes. In the FCIL settings, the proposed LGA model is efficient to tackle local and global catastrophic forgetting on old categories.

\begin{figure*}[t]
	\centering
	\includegraphics[width=520pt, height=115pt]
	{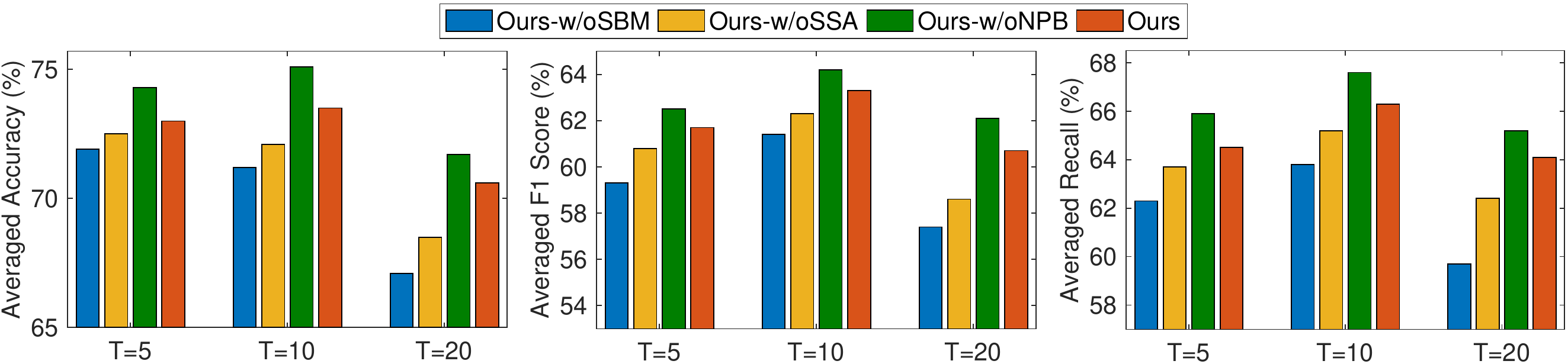}
	\vspace{-20pt}
	\caption{Effectiveness analysis of the proposed proxy server in terms of averaged accuracy, F1 score and recall on CIFAR-100 dataset \cite{krizhevsky2009learning}. }
	\label{fig: analysis_proxy_server_cifar100}
\end{figure*}

\begin{figure*}[t]
	\centering
	\includegraphics[width=520pt, height=115pt]
	{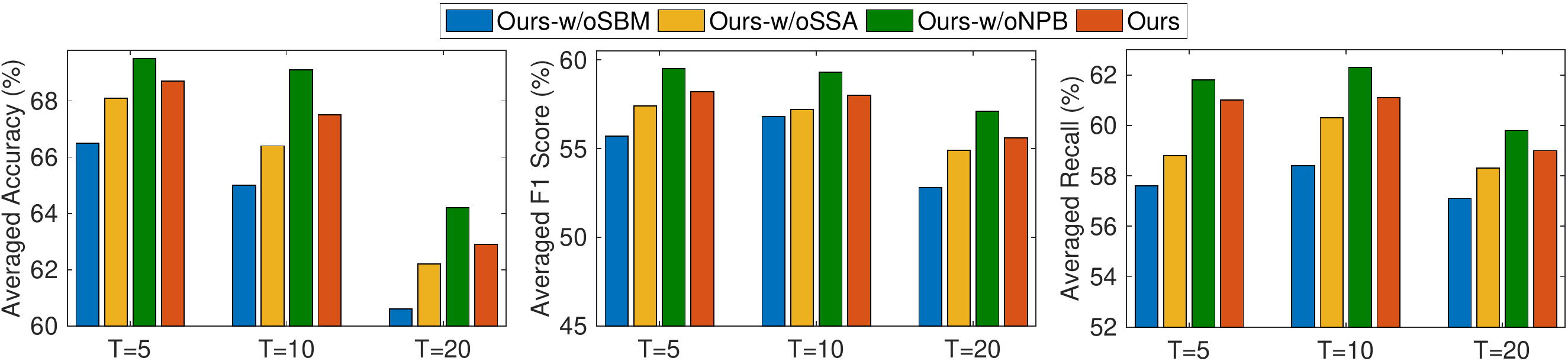}
	\vspace{-20pt}
	\caption{Effectiveness analysis of the proposed proxy server in terms of averaged accuracy, F1 score and recall on MiniImageNet dataset \cite{NIPS2016_90e13578}. }
	\label{fig: analysis_proxy_server_imagenet_subnet100}
\end{figure*}

\begin{table*}[t]
	\centering
	\setlength{\tabcolsep}{1.3mm}
	\caption{Investigation of exemplar memory $\mathcal{M}_l$ in terms of accuracy, F1 score and recall on CIFAR-100 \cite{krizhevsky2009learning} when the number of learning tasks $T=\{5, 10\}$. }
	\scalebox{0.999}{
		\begin{tabular}{c|c|ccccc>{\columncolor{lightgray}}c|>{\columncolor{lightgray}}c|cccccccccc|>{\columncolor{lightgray}}c|>{\columncolor{lightgray}}c}
			\toprule
			\multirow{2}{*}{} & \multirow{2}{*}{$\mathcal{M}_l$} & \multicolumn{7}{c|}{$T=5$} & \multicolumn{12}{c}{$T=10$} \\
			& & 20 & 40 & 60 & 80 & 100 & Avg. (\%) & Imp. (\%) & 10 & 20 & 30 & 40 & 50 & 60 & 70 & 80 & 90 & 100 & Avg. (\%) & Imp. (\%) \\
			
			\midrule
			\multirow{4}{*}{\rotatebox{90}{Accuracy}}& 
			500 & 82.5& 65.8& 63.2& 57.0& 49.4 & 63.6 & $\Uparrow$ 9.4 & 89.6& 80.1& 74.3& 69.1& 63.5& 61.6& 57.4& 54.6& 52.1& 50.1 & 65.2 & $\Uparrow$ 8.3 \\ 
			& 1000 & 83.1& 74.0& 65.0& 60.4& 55.3 & 67.6 & $\Uparrow$ 5.4 & 88.7& 80.8& 77.0& 74.0& 69.9& 67.9& 61.4& 59.6& 56.5& 55.6 & 69.1 & $\Uparrow$ 4.4 \\ 
			& 1500 &82.8& 71.8& 68.0& 61.7& 57.1& 68.3 & $\Uparrow$ 4.7 & 88.0& \textcolor[rgb]{0.698,0.133,0.133}{\textbf{83.3}}& 78.6& 75.8& 71.9& 68.7& 65.6& 62.9& 59.9& 59.5 & 71.4 & $\Uparrow$ 2.1 \\ 
			& 2000 & \textcolor[rgb]{0.698,0.133,0.133}{\textbf{83.3}}& \textcolor[rgb]{0.698,0.133,0.133}{\textbf{77.3}}& \textcolor[rgb]{0.698,0.133,0.133}{\textbf{72.8}}& \textcolor[rgb]{0.698,0.133,0.133}{\textbf{67.8}}& \textcolor[rgb]{0.698,0.133,0.133}{\textbf{63.7}} & \textcolor[rgb]{0.698,0.133,0.133}{\textbf{73.0}} & --- &\textcolor[rgb]{0.698,0.133,0.133}{\textbf{89.7}}& 83.2& \textcolor[rgb]{0.698,0.133,0.133}{\textbf{79.3}}& \textcolor[rgb]{0.698,0.133,0.133}{\textbf{76.1}}& \textcolor[rgb]{0.698,0.133,0.133}{\textbf{72.9}}& \textcolor[rgb]{0.698,0.133,0.133}{\textbf{71.7}}& \textcolor[rgb]{0.698,0.133,0.133}{\textbf{68.4}}& \textcolor[rgb]{0.698,0.133,0.133}{\textbf{65.7}}& \textcolor[rgb]{0.698,0.133,0.133}{\textbf{64.7}}& \textcolor[rgb]{0.698,0.133,0.133}{\textbf{62.9}} & \textcolor[rgb]{0.698,0.133,0.133}{\textbf{73.5}} & --- \\ 
			
			\midrule
			\multirow{4}{*}{\rotatebox{90}{F1 Score}}
			& 500 & 81.2& 58.2& 52.1& 41.5& 31.3 & 52.9 & $\Uparrow$ 8.8 & 89.3& 78.9& 70.8& 62.6& 55.2& 49.8& 44.9& 40.4& 36.0& 33.5 & 56.1 & $\Uparrow$ 7.8\\ 
			& 1000 & 81.7& 68.3& 53.7& 45.1& 36.8 & 57.1 & $\Uparrow$ 4.6 &88.3& 79.8& 73.5& 68.9& 61.6& 56.5& 47.6& 44.3& 39.2& 37.0 & 59.7 & $\Uparrow$ 4.2 \\ 
			& 1500 &81.6& 65.0& 56.7& 46.2& 38.0 & 57.5 & $\Uparrow$ 4.2 & 87.7& 81.4& 74.9& 69.6& 63.7&57.3& 51.6& 47.0& 42.2& 40.3 & 61.6 & $\Uparrow$ 2.3 \\ 
			& 2000 & \textcolor[rgb]{0.698,0.133,0.133}{\textbf{82.2}}& \textcolor[rgb]{0.698,0.133,0.133}{\textbf{71.8}}& \textcolor[rgb]{0.698,0.133,0.133}{\textbf{60.5}}& \textcolor[rgb]{0.698,0.133,0.133}{\textbf{50.6}} &\textcolor[rgb]{0.698,0.133,0.133}{\textbf{43.2}} & \textcolor[rgb]{0.698,0.133,0.133}{\textbf{61.7}} & --- &\textcolor[rgb]{0.698,0.133,0.133}{\textbf{89.4}}& \textcolor[rgb]{0.698,0.133,0.133}{\textbf{81.7}}& \textcolor[rgb]{0.698,0.133,0.133}{\textbf{76.1}}& \textcolor[rgb]{0.698,0.133,0.133}{\textbf{70.9}}& \textcolor[rgb]{0.698,0.133,0.133}{\textbf{65.0}}& \textcolor[rgb]{0.698,0.133,0.133}{\textbf{61.0}}& \textcolor[rgb]{0.698,0.133,0.133}{\textbf{55.5}}& \textcolor[rgb]{0.698,0.133,0.133}{\textbf{49.9}}& \textcolor[rgb]{0.698,0.133,0.133}{\textbf{46.4}}& \textcolor[rgb]{0.698,0.133,0.133}{\textbf{42.9}}& \textcolor[rgb]{0.698,0.133,0.133}{\textbf{63.9}}& ---  \\  
			
			\midrule
			\multirow{4}{*}{\rotatebox{90}{Recall}}
			& 500 & 82.7& 60.9& 54.9& 43.6& 32.6 & 54.9 & $\Uparrow$ 8.8 & \textcolor[rgb]{0.698,0.133,0.133}{\textbf{90.2}}& 81.9& 74.1& 66.6& 60.0& 53.7& 48.2& 43.8& 38.5& 35.3 & 59.2 & $\Uparrow$ 7.1 \\ 
			& 1000 &82.7& 71.8& 57.6& 47.6& 38.7 & 59.7 & $\Uparrow$ 4.0 & 89.2& 81.5& 76.2& 72.2& 65.5& 60.1& 50.9& 47.1& 41.6& 38.9 & 62.3 & $\Uparrow$ 4.0 \\ 
			& 1500 & 82.7& 68.5& 59.5& 49.0& 39.7 & 59.9 & $\Uparrow$ 3.8 &88.5& 83.2& 77.5& 72.8& 66.9& 60.4& 54.0& 49.4& 44.3& 42.2 & 63.9 & $\Uparrow$ 2.4 \\ 
			& 2000 & \textcolor[rgb]{0.698,0.133,0.133}{\textbf{83.6}} & \textcolor[rgb]{0.698,0.133,0.133}{\textbf{74.5}}& \textcolor[rgb]{0.698,0.133,0.133}{\textbf{63.1}}& \textcolor[rgb]{0.698,0.133,0.133}{\textbf{53.4}}& \textcolor[rgb]{0.698,0.133,0.133}{\textbf{44.1}}& \textcolor[rgb]{0.698,0.133,0.133}{\textbf{63.7}} & --- & 89.9& \textcolor[rgb]{0.698,0.133,0.133}{\textbf{83.4}}& \textcolor[rgb]{0.698,0.133,0.133}{\textbf{78.3}}& \textcolor[rgb]{0.698,0.133,0.133}{\textbf{74.1}}& \textcolor[rgb]{0.698,0.133,0.133}{\textbf{68.1}}& \textcolor[rgb]{0.698,0.133,0.133}{\textbf{64.1}}& \textcolor[rgb]{0.698,0.133,0.133}{\textbf{58.3}}& \textcolor[rgb]{0.698,0.133,0.133}{\textbf{52.6}}& \textcolor[rgb]{0.698,0.133,0.133}{\textbf{48.8}}& \textcolor[rgb]{0.698,0.133,0.133}{\textbf{45.0}} & \textcolor[rgb]{0.698,0.133,0.133}{\textbf{66.3}} & --- \\  
			\bottomrule
		\end{tabular}
	} 	
	\label{tab: memory_cifar100_tasks5/10}
\end{table*}

\subsection{Analysis of Heterogeneous Class Distribution}
In order to discuss how heterogeneous class distribution across local clients affects performance, as presented in Table~\ref{tab: heterogeneous_cifar100_tasks10}, we conduct extensive experiments in terms of accuracy and F1 score on CIFAR-100 dataset \cite{krizhevsky2009learning}. Specifically, we consider different settings of data heterogeneity via randomly assigning $\{40\%, 50\%, 60\%\}$ categories from the label space of current learning task to local clients. From Table~\ref{tab: heterogeneous_cifar100_tasks10}, we conclude that the proposed LGA model achieves significant performance improvement about $3.6\%\sim6.6\%$ in terms of averaged accuracy and F1 score under different settings of heterogeneous class distributions across local clients, when compared with our conference version (\emph{i.e.}, GLFC \cite{Dong_2022_CVPR}). It validates the effectiveness of our model to tackle local and global forgetting on old categories brought by heterogeneous class imbalance across local clients. Although the increase of data heterogeneity at local side degrades the performance of proposed model, our LGA model still achieves better performance to tackle large data heterogeneity brought by Non-IID class imbalance across local clients when compared with GLFC \cite{Dong_2022_CVPR}. Table~\ref{tab: heterogeneous_cifar100_tasks10} also illustrates superiority and effectiveness of the ameliorated proxy server to tackle global catastrophic forgetting on old categories.

\begin{figure*}[t]
	\centering
	\includegraphics[width=520pt, height=110pt]
	{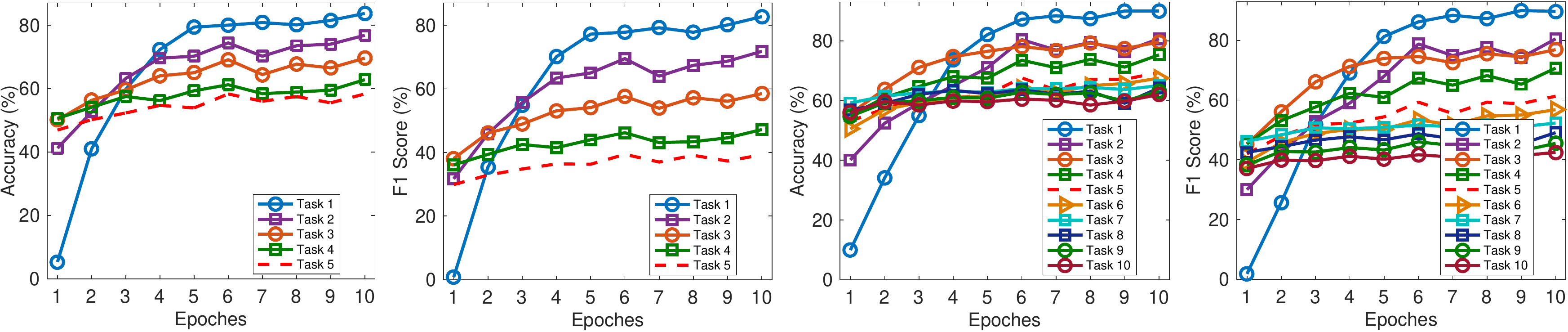}
	\vspace{-20pt}
	\caption{Convergence analysis of our LGA model in terms of accuracy and F1 score on CIFAR-100 dataset \cite{krizhevsky2009learning} when $T=5$ (left) and $T=10$ (right). }
	\label{fig: convergence_analysis}
\end{figure*}

\subsection{Analysis of Proxy Server}
In this subsection, we investigate the effectiveness of proxy server to surmount global catastrophic forgetting on old classes by analyzing the performance of each proposed strategy on CIFAR-100 \cite{krizhevsky2009learning} and MiniImageNet \cite{NIPS2016_90e13578} datasets, as shown in Figs.~\ref{fig: analysis_proxy_server_cifar100} and \ref{fig: analysis_proxy_server_imagenet_subnet100}. We denote the performance (averaged accuracy, F1 score and recall) of proposed proxy server without using the selection of best old model, self-supervised prototype augmentation and noisy perturbation in latent feature space as Ours-w/oSBM, Ours-w/oSSA and Ours-w/oNPB respectively. As presented in Figs.~\ref{fig: analysis_proxy_server_cifar100} and \ref{fig: analysis_proxy_server_imagenet_subnet100}, when compared with Ours, Ours-w/oSBM has a large performance degradation, indicating the importance of selecting the best old model to tackle global forgetting on old categories brought by Non-IID class imbalance across clients. The result degradation of Ours-w/oSSA illustrates the superiority to augment prototype images under a self-supervised manner, which is useful for the selection of old model to improve distillation gains of the category gradient-induced semantic distillation loss $\mathcal{L}_{\mr{SD}}$. Furthermore, Ours-w/oNPB shows comparable performance with Ours, since noisy perturbation is added to latent feature space for privacy preservation instead of performance improvement. As shown in Fig.~\ref{fig: prototype_samples}, we visualize some original prototype images and their reconstructed prototype images via proxy server to better understand privacy protection of local clients.

\textbf{Communication Cost:} As shown in Table~\ref{tab: communication_cost}, the communication cost of proxy server covers only one round of 4-layer $\Gamma$ gradients (16.5MB) of prototype exemplars for each training step. Compared to the communication cost (4536.8MB) of traditional FL \cite{DBLP:journals/corr/McMahanMRA16}, we have 0.364\% additional cost (16.5MB) since local clients transmit only one representative prototype of each new class to proxy server. Such low communication cost is acceptable in real-world applications when we have large performance improvement over existing baseline methods (see Tables~\ref{tab: exp_all_datasets_task5}$\sim$\ref{tab: exp_tinyimageNet_task20}). Moreover, our proposed proxy server can effectively save memory storage of local clients via selecting the best old model.

\subsection{Analysis of Exemplar Memory Size}
To evaluate the effects of local exemplar memory $\mathcal{M}_l=\{500, 1000, 1500, 2000\}$ on the performance of our model, as presented in Table~\ref{tab: memory_cifar100_tasks5/10}, we conduct extensive experiments on CIFAR-100 \cite{krizhevsky2009learning} when setting the number of consecutive learning tasks $T=\{5, 10\}$. The experimental results presented in Table~\ref{tab: memory_cifar100_tasks5/10} indicate that a larger exemplar memory $\mathcal{M}_l$ can effectively improve the performance of proposed LGA model. It validates the superiority of our model to surmount local and global catastrophic forgetting on old categories when local clients use a larger memory overhead to replay more training images of old categories. Moreover, Table~\ref{tab: memory_cifar100_tasks5/10} illustrates that it is essential for local clients to update exemplar memory $\mathcal{M}_l$ and store old categories, after detecting new categories via task transition detection. In the FCIL settings, the exemplar memory $\mathcal{M}_l$ at local side can effectively tackle local forgetting on old categories, further promoting the performance of global anti-forgetting.

\begin{table}[t]
\centering
\setlength{\tabcolsep}{2.3mm}
\caption{Analysis about varying degrees of incremental tasks on CIFAR-100 \cite{krizhevsky2009learning} when $T=\{5, 10, 20\}$. } 
\scalebox{0.995}{
    \begin{tabular}{c|c|ccc}
        \toprule
       Settings & Methods & $T=5$ & $10$ & $20$ \\
        \midrule
        \multirow{2}{*}{Task Order \#1} & GLFC \cite{Dong_2022_CVPR} & 68.3 & 66.9 & 66.4 \\
        & \textbf{Ours (LGA)} & 73.0 & 73.5 & 70.6 \\
        \midrule
        \multirow{2}{*}{Task Order \#2} & GLFC \cite{Dong_2022_CVPR} & 67.8 & 67.3 & 66.7 \\
        & \textbf{Ours (LGA)} & 73.3 & 73.7 & 72.1 \\
        \midrule
        \multirow{2}{*}{Task Order \#3} & GLFC \cite{Dong_2022_CVPR} & 68.2 & 67.8 & 67.3 \\
        & \textbf{Ours (LGA)} & 73.5 & 73.2 & 71.4 \\
        \bottomrule
\end{tabular}	} 	
\label{tab: varying_incremental_task}
\end{table}

\subsection{Analysis of Varying Incremental Tasks}

As shown in Table~\ref{tab: varying_incremental_task}, we conduct experiments on CIFAR-100 \cite{krizhevsky2009learning} to investigate whether varying degrees of incremental tasks can affect the performance (accuracy) of our LGA model. In Table~\ref{tab: varying_incremental_task}, the task order \#1 denotes baseline task settings proposed by iCaRL \cite{Rebuffi_2017_CVPR} and GLFC \cite{Dong_2022_CVPR}. As for the settings of task order \#2 and task order \#3, we randomly assign different classes from CIFAR-100 \cite{krizhevsky2009learning} to each incremental task. It ensures the randomness of task orders and classes within each incremental task. We can observe that the performance of our model decreases slightly for larger number of incremental tasks ($T=20$), but still outperforms baseline method GLFC \cite{Dong_2022_CVPR} by 4.1\%$\sim$6.6\% averaged accuracy. Moreover, our model has little performance variance for different task orders, which shows robustness of our model to tackle different FCIL settings. From the comparisons between $T=5$ and $T=10$, we conclude that the number of classes learned in each task has negligible influence on performance, while a larger number of incremental tasks ($T=20$) may decrease the performance slightly.

\subsection{Convergence Analysis}
As shown in Fig.~\ref{fig: convergence_analysis}, we investigate the convergence analysis of our model in terms of accuracy and F1 score on CIFAR-100 \cite{krizhevsky2009learning} when setting $T=\{5, 10\}$. From the depicted curves in Fig.~\ref{fig: convergence_analysis}, we observe the proposed LGA model could efficiently converge to a stable performance after a few iteration epoches. It also validates that our LGA model could effectively learn a global class-incremental model to tackle the FCIL problem under the privacy preservation. Both local and global forgetting on old categories can be alleviated via the category-balanced gradient-adaptive compensation loss and category gradient-induced semantic distillation loss, when the best old global model is selected via proxy server.

\section{Conclusion}
In this paper, we focus on addressing a real-world FL challenge named Federated Class-Incremental Learning (FCIL), and propose a Local-Global Anti-forgetting (LGA) model to overcome local and global forgetting on old categories under the FCIL settings. To be specific, a category-balanced gradient-adaptive compensation loss and a category gradient-induced semantic distillation loss are designed to tackle local forgetting brought by local clients' class imbalance, by compensating heterogeneous forgetting speeds of old categories while distilling consistent class relations within different tasks. Furthermore, considering addressing global forgetting caused by Non-IID class imbalance across local clients, we develop a proxy server to reconstruct perturbed images of new categories via prototype gradient communication under privacy protection, and augment them via self-supervised prototype augmentation to select the best old global model from a global perspective. We illustrate the superiority of proposed LGA model against baseline methods via qualitative experiments on several representative datasets.

In the future, we will introduce theoretical analysis to guarantee the stable convergence of our LGA model from a theory perspective, and extend the proposed LGA model into other challenging federated learning based vision tasks such as semantic segmentation, visual tracking and object detection. Moreover, we will explore how the global server can automatically detect the number of learned classes in each incremental learning task without using human prior.

\ifCLASSOPTIONcaptionsoff
  \newpage
\fi

\bibliographystyle{plain}
\bibliography{TPAMI2023_FinalVersion}

\begin{IEEEbiography}[{\includegraphics[width=1in,height=1.25in,clip,keepaspectratio]{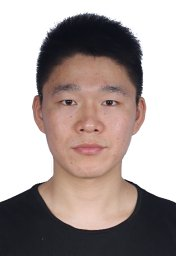}}]{Jiahua Dong} is currently a PhD candidate in the State Key Laboratory of Robotics, Shenyang Institute of Automation, Chinese Academy of Sciences. He received the B.S. degree from Jilin University in 2017. He visited ETH Zurich, Switzerland from Apr. 2022 to Aug. 2022, and Max Planck Institute for Informatics, Germany from Sep. 2022 to Jan. 2023. His current research interests include transfer learning, class-incremental learning, federated learning and medical image processing. 
\end{IEEEbiography}

\begin{IEEEbiography}[{\includegraphics[width=1in,height=1.25in,clip,keepaspectratio]{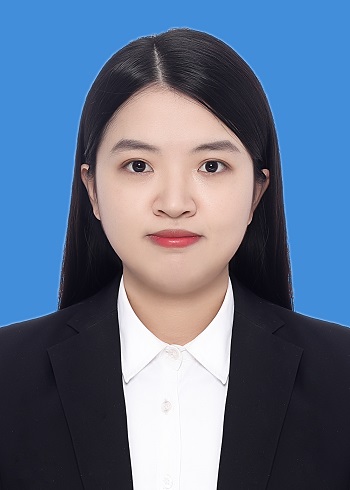}}]{Hongliu Li} is currently a postdoctoral fellow in the Department of Civil and Environmental Engineering, Hong Kong Polytechnic University. She received the B.S. degree from University of Petroleum of China in 2017, and the Ph.D degree from University of Science and Technology of China, and the Ph.D. degree from City University of Hong Kong in 2022. Her current research interests include computer vision, machine learning, pedestrian and evacuation dynamics. 
\end{IEEEbiography}

\begin{IEEEbiography}[{\includegraphics[width=1in,height=1.25in,clip,keepaspectratio]{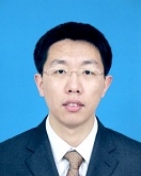}}]{Yang Cong} (Senior Member, IEEE) received the B.Sc. degree from Northeast University in 2004 and the Ph.D. degree from the State Key Laboratory of Robotics, Chinese Academy of Sciences, in 2009. From 2009 to 2011, he was a Research Fellow with the National University of Singapore (NUS) and Nanyang Technological University (NTU). He was a Visiting Scholar with the University of Rochester. He was the professor until 2023 with Shenyang Institute of Automation, Chinese Academy of Sciences. He is currently the full professor with South China University of Technology. He has authored over 80 technical articles. His current research interests include robot, computer vision, machine learning, multimedia, medical imaging and data mining. He has served on the editorial board of the several joural papers. He was a senior member of IEEE since 2015. 
\end{IEEEbiography}

\begin{IEEEbiography}[{\includegraphics[width=1in,height=1.25in,clip,keepaspectratio]{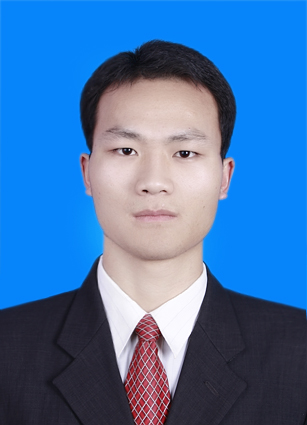}}]{Gan Sun} (S'19-M'20) is an associate professor in State Key Laboratory of Robotics, Shenyang Institute of Automation, Chinese Academy of Sciences. He received the B.S. degree from Shandong Agricultural University in 2013, the Ph.D. degree from State Key Laboratory of Robotics, Shenyang Institute of Automation, Chinese Academy of Sciences in 2020, and has been visiting Northeastern University from April 2018 to May 2019, Massachusetts Institute of Technology from June 2019 to November 2019. His current research interests include lifelong machine learning, multitask learning, medical data analysis, deep learning and 3D computer vision.
\end{IEEEbiography}

\begin{IEEEbiography}[{\includegraphics[width=1in,height=1.25in,clip,keepaspectratio]{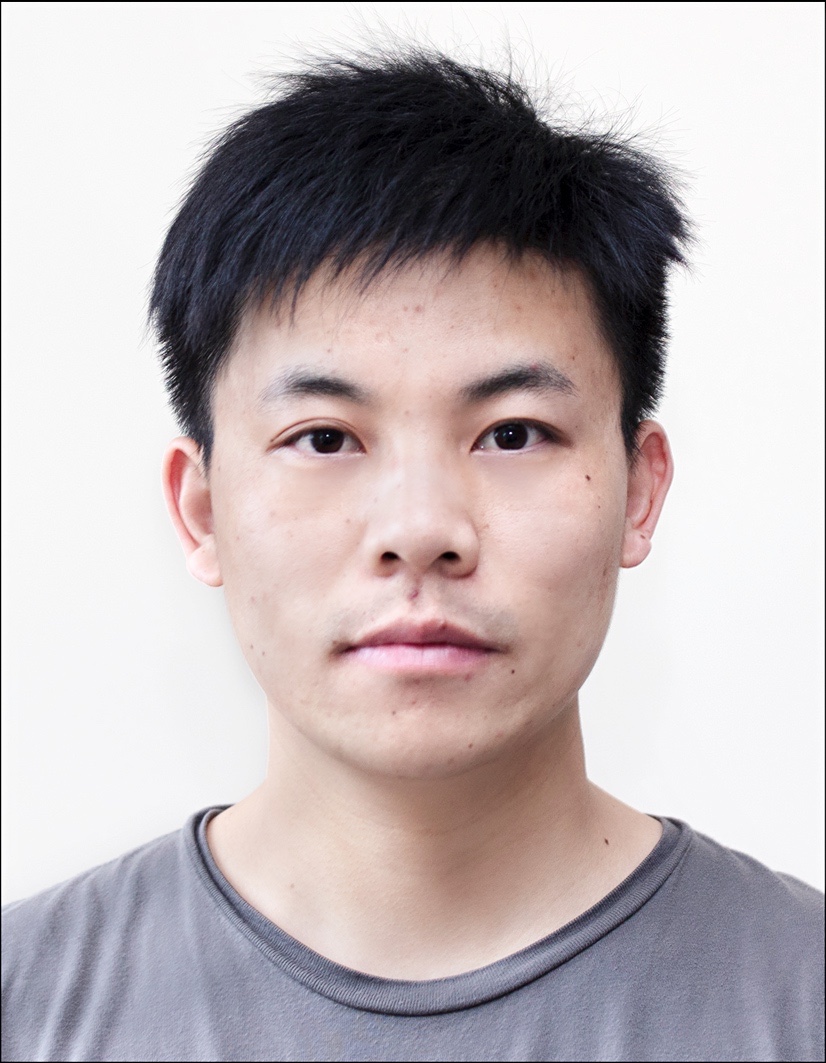}}]{Yulun Zhang} is a postdoctoral researcher at Computer Vision Lab, ETH Zürich, Switzerland. He obtained the Ph.D. degree from the Department of ECE, Northeastern University, USA, in 2021. He also worked as a research fellow in Harvard University. Before that, he received the B.E. degree from the School of Electronic Engineering, Xidian University, China, in 2013 and the M.E. degree from the Department of Automation, Tsinghua University, China, in 2017. His research interests include image/video restoration and synthesis, biomedical image analysis, model compression, and computational imaging.
\end{IEEEbiography}

\begin{IEEEbiography}[{\includegraphics[width=1in,height=1.25in,clip,keepaspectratio]{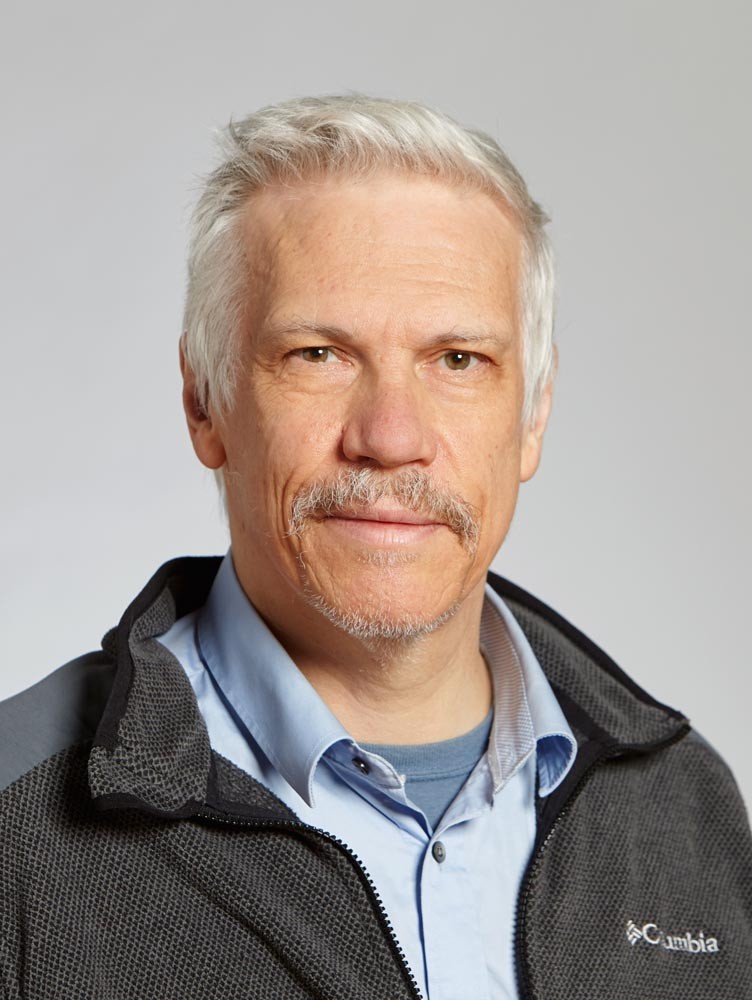}}]{Luc Van Gool} received the degree in electromechanical engineering from the Katholieke Universiteit Leuven, Leuven, Belgium, in 1981. He is currently a professor at the Katholieke Universiteit Leuven, Belgium, and the ETH, Zurich, Switzerland. He leads computer vision research with both places, where he also teaches computer vision. He has authored more than 200 papers in this field. His research interests mainly include 3D reconstruction and modeling, object recognition, tracking, and gesture analysis. He was a program committee member of several major computer vision conferences. He was the recipient of several best paper awards. He is a cofounder of five spin-off companies.
\end{IEEEbiography}

\end{document}